\documentclass{article}

\PassOptionsToPackage{numbers,compress}{natbib}
\PassOptionsToPackage{table,dvipsnames}{xcolor}

 \usepackage[eandd, final]{neurips_2026}

\usepackage[utf8]{inputenc} 
\usepackage[T1]{fontenc}    
\usepackage{url}            
\usepackage{booktabs}       
\usepackage{amsfonts}       
\usepackage{amsmath}        
\usepackage{nicefrac}       
\usepackage{microtype}      
\usepackage{xcolor}         

\usepackage{graphicx}
\usepackage{svg}
\usepackage{tabularx}
\usepackage{makecell}
\usepackage{amssymb}
\usepackage{array}
\usepackage{enumitem}
\usepackage[detect-all]{siunitx}
\usepackage{multirow}
\usepackage{placeins}

\definecolor{BestCell}{HTML}{DCEBFF}
\definecolor{SecondCell}{HTML}{F0F6FF}
\definecolor{GroupRow}{HTML}{F4F4F4}

\newcommand{\best}[1]{\cellcolor{BestCell}\textbf{#1}}
\newcommand{\second}[1]{\cellcolor{SecondCell}\underline{#1}}
\newcommand{\groupheader}[1]{%
\rowcolor{GroupRow}\multicolumn{12}{@{}l}{\textbf{\textit{#1}}}\\}

\definecolor{citeblue}{rgb}{0.21,0.49,0.74}
\definecolor{lightgray}{gray}{0.94}
\usepackage[breaklinks,colorlinks,citecolor=citeblue,linkcolor=purple,urlcolor=citeblue]{hyperref}

\title{RAVEN-Bench: A Paired EO--IR Video QA Benchmark for Aerial Multimodal Understanding}

\author{%
  Yu Hu\textsuperscript{1}\quad
  Jianyang Gu\textsuperscript{2}\quad
  Yue Cao\textsuperscript{1}\quad
  Hao Liu\textsuperscript{1}\\
  \textbf{Kangnan Wang\textsuperscript{1}}\quad
  \textbf{Jozsef Hamari\textsuperscript{3}}\quad
  \textbf{Zheng Liu\textsuperscript{1$\dagger$}}\quad
  \textbf{Mohsen Zardadi\textsuperscript{3$\dagger$}}\\
  \textsuperscript{1}The University of British Columbia, Okanagan, Kelowna, BC, Canada\\
  \textsuperscript{2}The Ohio State University, Columbus, OH, USA\\
  \textsuperscript{3}TerraSense Analytics, Kelowna, BC, Canada\\
  \texttt{\{yu.hu, yue.cao, hao.liu, kangnan.wang, zheng.liu\}@ubc.ca}\\
  \texttt{gu.1220@osu.edu}\quad
  \texttt{\{jozsef.hamari, mohsen.zardadi\}@terrasense.ca}
}

\begin{document}

\maketitle
{\renewcommand{\thefootnote}{}\footnotetext{$^\dagger$Corresponding authors.}}


\begin{abstract}
Multimodal large language models (MLLMs) have advanced rapidly in image and video understanding, yet their ability to reason over paired aerial electro-optical and infrared (EO--IR) videos remains poorly understood. Such video poses a harder test than standard RGB video QA: targets are often small, evidence may appear only briefly, viewpoints change with the platform, and useful cues may appear in only one stream. We introduce \textbf{RAVEN-Bench} (\textbf{R}easoning over \textbf{A}erial \textbf{V}ideos with \textbf{E}O--IR alig\textbf{N}ment), a paired EO--IR video QA benchmark for low-altitude aerial multimodal understanding. RAVEN-Bench contains 100 temporally aligned EO--IR video pairs and 1,200 human-authored four-way multiple-choice questions, each independently audited for answer keys, modality requirements, evidence intervals, and group validity. It supports EO-only, IR-only, and paired EO+IR evaluation, and uses hierarchical capability levels and structured question groups to diagnose paired-input gain beyond the stronger unimodal condition, unimodal reliance, target consistency, and evidence-chain coherence. We evaluate frontier, open-weight, and remote-sensing MLLMs, showing that aggregate accuracy can overstate evidence-supported EO--IR reasoning. RAVEN-Bench provides controlled modality conditions and group-level diagnostics for auditing whether progress on RGB video benchmarks transfers to aerial cross-spectral understanding. Code and data are available at \href{https://github.com/yuhu990424/RAVEN-Bench}{\nolinkurl{github.com/yuhu990424/RAVEN-Bench}} and \href{https://huggingface.co/datasets/TerraSenseAnalytics/RAVEN-Bench}{\nolinkurl{huggingface.co/datasets/TerraSenseAnalytics/RAVEN-Bench}}.
\end{abstract}

\section{Introduction}

Multimodal large language models (MLLMs) have made rapid progress in image and video understanding. 
Recent systems can describe dynamic scenes, follow visual instructions, and answer questions grounded in temporal evidence~\citep{bai2025qwen25vl,chen2024internvl25,lin2023videollava,zhang2024llavavideo}. 
Much of this progress has been measured on video benchmarks such as Video-MME~\citep{fu2024videomme}, MLVU~\citep{zhou2025mlvu}, and LVBench~\citep{wang2025lvbench}. 
These benchmarks mainly use RGB natural videos, including web videos, movies, egocentric clips, instructional scenes, surveillance footage, and daily activities. 
Aerial electro-optical and infrared (EO--IR) video poses a different question, since the evidence can be small, brief, and tied to how each sensor responds to the scene.

Low-altitude EO--IR video often looks understandable at the scene level while remaining difficult at the target level. 
Aircraft, vessels, and vehicles may occupy only a small part of the frame, appear briefly under platform motion, or become ambiguous under occlusion, shadow, haze, low contrast, and background clutter. 
EO streams preserve shape, texture, color, and scene layout. 
IR streams expose thermal contrast that may become useful when visible-spectrum evidence is weak. 
For example, a vehicle may remain visually identifiable in EO across viewpoint changes, while its IR contrast varies as it moves between cooler background regions and warmer pavement. 
These conditions make aggregate video-QA accuracy a limited diagnostic signal. 
A model may answer an isolated question correctly while losing the same target in a related question. 
It may choose a plausible event-level answer without recognizing the intermediate visual evidence that supports it. 
It may also appear to benefit from EO+IR input while relying mainly on the stronger unimodal stream. 
Aerial EO--IR evaluation therefore needs controlled modality settings, temporally grounded questions, and group-level measures of target consistency and evidence-chain coherence.
Before MLLMs are applied to non-RGB aerial video, researchers should be able to audit whether a model uses the sensor that carries the evidence, keeps the same target, and supports event-level answers with earlier evidence; RAVEN-Bench is built for this research audit, not for operational deployment.

Existing resources cover parts of this problem, but the paired EO--IR video reasoning setting remains largely untested. 
General video MLLM benchmarks evaluate temporal reasoning mainly in RGB natural videos. 
Remote-sensing vision-language benchmarks study captioning, grounding, referring, and VQA, with most evaluations centered on images~\citep{kuckreja2024geochat,li2024vrsbench,zhang2024earthgpt}. 
EO--IR and multispectral aerial datasets show the value of cross-spectral sensing, with tasks usually framed as detection, tracking, segmentation, or sensor-fusion perception~\citep{wu2014thermal,qin2025must,manjunath2024indraeye}. 
The missing setting is language-grounded reasoning over paired EO--IR aerial videos, where models must identify target-centric evidence, follow how it changes over time, and decide when EO, IR, or both streams are needed.

\begin{figure*}[t]
    \centering
    \includegraphics[width=\linewidth]{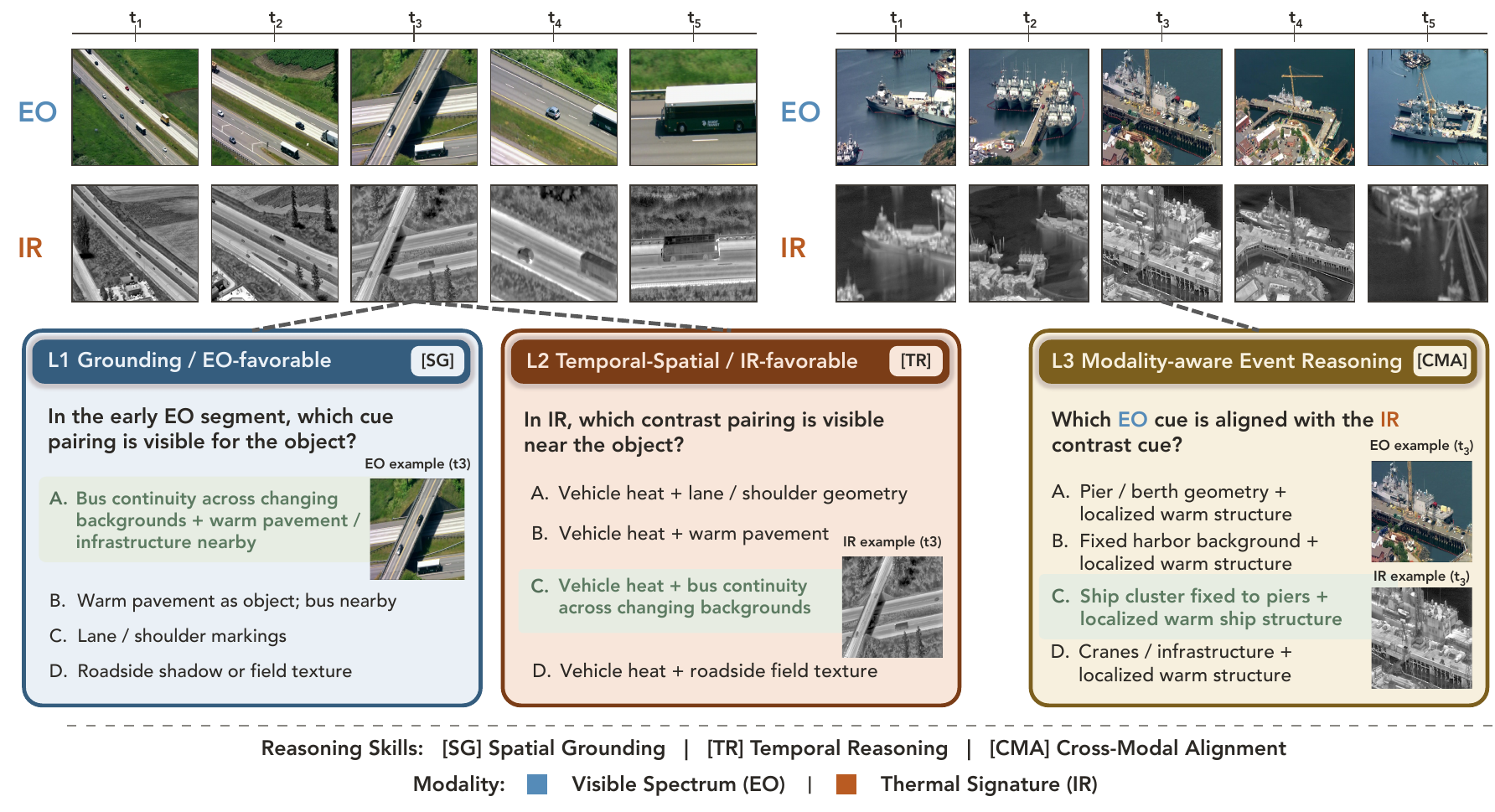}
\caption{
    Overview of a \textbf{RAVEN-Bench} sample with synchronized EO--IR observations and representative L1--L3 four-way QA items. Labels indicate capability level and item-level modality requirement for analysis only, and are not provided as model input.
    }
    \label{fig:raven_overview}
\end{figure*}

We introduce \textbf{RAVEN-Bench} (\textbf{R}easoning over \textbf{A}erial \textbf{V}ideos with \textbf{E}O--IR alig\textbf{N}ment), a curated diagnostic benchmark for paired EO--IR aerial video QA. 
RAVEN-Bench contains 100 EO--IR video pairs and 1,200 human-authored four-way multiple-choice QA items, with twelve questions per video pair. Every item is independently audited. 
Each sample centers on one target-centric aerial event and supports EO-only, IR-only, and paired EO+IR evaluation. 
As illustrated in \autoref{fig:raven_overview}, a sample pairs synchronized EO and IR video observations with questions that move from local grounding to temporal-spatial reasoning and modality-aware event reasoning.

RAVEN-Bench focuses on language-grounded reasoning over paired EO--IR video evidence and provides no annotations or metrics for detection, tracking, segmentation, geolocation, targeting, or pixel-level registration; users with access could still create new annotations from the videos (Appendix~\ref{app:responsible_release}). 
The questions span three capability levels: grounding and cross-spectral evidence aggregation, temporal/spatial reasoning under modality shift, and modality-aware event reasoning. 
Related questions are grouped so that the same target or event is revisited across observations and ordered reasoning steps. 
This structure allows RAVEN-Bench to measure answer correctness together with unimodal reliance, paired-input gain beyond the stronger unimodal condition, stable target interpretation, and evidence-supported event reasoning. 

Our contributions are summarized as follows:
\begin{itemize}[leftmargin=18pt,nosep]
    \item We introduce \textbf{RAVEN-Bench}, a curated paired EO--IR aerial video QA benchmark with 100 EO--IR video pairs and 1,200 independently audited four-way QA items for target-centric aerial multimodal understanding.
    
    \item We design an evidence-conditioned evaluation protocol with EO-only, IR-only, and paired EO+IR settings, enabling controlled evidence-condition slices and measurement of paired-input gain beyond the stronger unimodal input.
    
    \item We develop a diagnostic structure with capability-balanced scoring, consistency groups, coherence groups, and cross-modal gain (\(G_{\mathrm{cm}}\), paired-input gain beyond the stronger unimodal condition), supported by audited annotations and evaluations of 26 MLLMs, including three remote-sensing VLMs.
\end{itemize}

\section{Related Work}

We compare prior resources by input format, sensor pairing, labels, correspondence controls, task, and scale (\autoref{tab:benchmark_positioning}).

\paragraph{Video understanding benchmarks.}
Benchmarks such as MVBench~\citep{li2024mvbench}, Video-MME~\citep{fu2024videomme}, MLVU~\citep{zhou2025mlvu}, and LVBench~\citep{wang2025lvbench} have been useful for testing temporal reasoning in RGB natural videos.
Video-MME-v2~\citep{he2026videommev2} additionally studies agreement across related questions and multi-step reasoning chains.
These benchmarks use RGB video only; RAVEN-Bench evaluates such agreement and reasoning chains in paired EO--IR aerial video, with controlled input conditions and correspondence tests.

\paragraph{Remote-sensing VLMs and benchmarks.}
GeoChat~\citep{kuckreja2024geochat}, EarthGPT~\citep{zhang2024earthgpt}, and VRSBench~\citep{li2024vrsbench} broadened language-grounded aerial understanding through remote-sensing captioning, grounding, and VQA.
SkyEyeGPT~\citep{zhan2024skyeyegpt} unifies image- and region-level remote-sensing tasks and also covers aerial video captioning. EarthDial~\citep{soni2025earthdial} handles multisensor, multitemporal, and multi-resolution Earth-observation imagery, including RGB, SAR, near-infrared, and infrared bands.
None of them evaluates paired EO--IR video under controlled EO-only, IR-only, and EO+IR inputs.

\paragraph{Optical--thermal VQA.}
Traffic-VQA~\citep{zhang2026trafficvqa} pairs 8,180 aligned optical and thermal UAV images with about 1.3M QA pairs. WildFireVQA~\citep{habibpour2026wildfirevqa} provides 6,097 aerial RGB--thermal samples with radiometric thermal data and 207,298 multiple-choice questions. Thermo-VL-Bench~\citep{thushara2026thermovl} evaluates pixel-aligned RGB--thermal image pairs under RGB-only, thermal-only, and RGB+thermal inputs.
These resources offer larger optical--thermal QA sets, calibrated thermal data, and paired-image evaluation, but all operate on still image pairs. RAVEN-Bench studies EO and IR evidence that changes within one event over time.

\paragraph{EO--IR and multispectral perception.}
Datasets such as TIV~\citep{wu2014thermal}, KAIST~\citep{hwang2015multispectral}, IndraEye~\citep{manjunath2024indraeye}, and MUST~\citep{qin2025must} cover EO--IR and multispectral sensing under difficult visibility conditions.
They target perception tasks such as detection, tracking, segmentation, and sensor fusion. RAVEN-Bench uses no spatial or trajectory labels, and its Consistency score does not measure tracking.

\autoref{tab:benchmark_positioning} summarizes these differences: RAVEN-Bench combines paired EO--IR video with controlled inputs, correspondence controls, and grouped measures of target consistency and evidence-chain coherence.

\providecommand{\tabyes}{\(\checkmark\)}
\providecommand{\tabno}{\textcolor{black!35}{--}}

\begin{table}[t]
\centering
\footnotesize
\setlength{\tabcolsep}{2pt}
\renewcommand{\arraystretch}{1.15}
\caption{Comparison of RAVEN-Bench with related benchmark families and the closest optical--thermal QA resources.}
\label{tab:benchmark_positioning}
\begin{tabular*}{\linewidth}{@{\extracolsep{\fill}}llllcll@{}}
\toprule
\textbf{Benchmark} & \textbf{Input} & \textbf{Sensor pairing} & \textbf{Labels} & \textbf{Corr.} & \textbf{Task} & \textbf{Scale} \\
\midrule
General video QA~\citep{li2024mvbench,fu2024videomme,he2026videommev2} & RGB video & -- & QA, groups$^{a}$ & -- & MCQ & -- \\
Long-video QA~\citep{wu2024longvideobench,zhou2025mlvu,wang2025lvbench,tan2025allvb} & RGB video & -- & QA & -- & VQA & -- \\
RS VLMs~\citep{kuckreja2024geochat,li2024vrsbench,zhan2024skyeyegpt,soni2025earthdial,wang2025xlrsbench,dang2025rshrbench} & RS image$^{b}$ & Multisensor & Cap., box, QA & -- & Cap., VQA & -- \\
EO--IR perception~\citep{manjunath2024indraeye,qin2025must} & EO--IR/MS & Paired & Box, mask, track & -- & Det., seg., track & -- \\
\midrule
Traffic-VQA~\citep{zhang2026trafficvqa} & UAV image & Optical--thermal & QA & -- & VQA & 8.2K / 1.3M \\
WildFireVQA~\citep{habibpour2026wildfirevqa} & UAV image & RGB--thermal & QA & -- & MCQ & 6.1K / 207K \\
Thermo-VL-Bench~\citep{thushara2026thermovl} & Image & RGB--thermal & QA & -- & Yes/no & 3.1K / 15K \\
\midrule
\textbf{RAVEN-Bench (ours)} & \textbf{EO--IR video} & \textbf{Synchronized} & \textbf{QA + metadata$^{c}$} & \tabyes & \textbf{MCQ, grouped} & \textbf{100 / 1,200} \\
\bottomrule
\end{tabular*}

\vspace{0.3em}
\begin{minipage}{\linewidth}
\scriptsize
Corr.: EO--IR correspondence controls (shuffled and temporally misaligned pairs). MS: multispectral. $^{a}$Video-MME-v2 only. $^{b}$Some add multisensor or aerial video inputs. $^{c}$Modality requirement, evidence interval, and group labels; no spatial labels. Scale: image or video pairs / QA items, listed for the closest paired optical--thermal resources.
\end{minipage}
\end{table}

\begin{figure}[t]
    \centering
    \includegraphics[width=\linewidth]{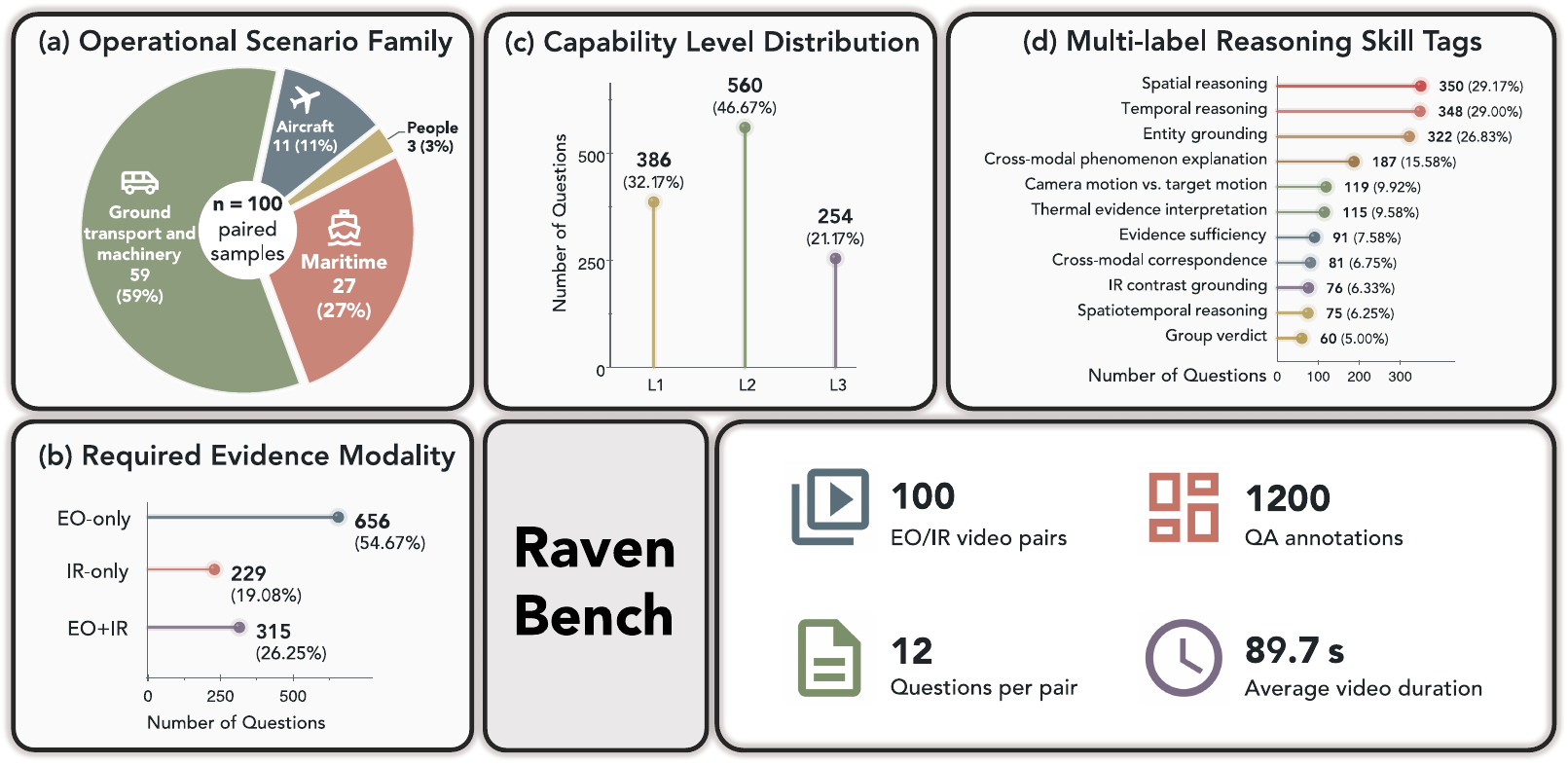}
    \caption{
    Dataset statistics for RAVEN-Bench. Panels show (a) scenario families over the 100 EO--IR pairs, (b) annotated evidence requirements, (c) capability levels, and (d) the 11 most frequent of 20 reasoning tags; (b)--(d) are computed over the 1,200 questions, and the mean duration over the 200 single-modality streams. Percentages in (d) need not sum to 100\%.
    }
    \label{fig:datastatistics}
\end{figure}

\section{RAVEN-Bench}
\label{sec:raven}

RAVEN-Bench contains 100 paired EO--IR aerial observations and 1,200 four-way QA items designed for target-centric, temporal, and modality-aware reasoning.
The benchmark task is language-grounded reasoning over paired EO--IR video evidence; RAVEN-Bench provides no annotations or metrics for detection, tracking, segmentation, geolocation, targeting, or pixel-level registration. 
Section~\ref{sec:task_design} defines the task design and diagnostic metrics.

\subsection{Video Source and Curation}
\label{sec:video_acquisition}

The videos come from an ongoing, weekly airborne collection program of TerraSense Analytics, which owns the footage and authorized its release for research evaluation; several authors are affiliated with it.
The footage was captured with a WESCAM MX-10 EO/IR gimbal on a Vulcanair P68C aircraft in western North America.
Exact dates, coordinates, flight paths, platform and sensor identifiers, and KLV and other operational metadata are withheld, and no third-party copyrighted video is included.
The gated release includes evaluation code, Croissant metadata, a dataset card, and research-use terms (Appendix~\ref{app:responsible_release}).

\paragraph{Scene families.}
The retained events cover four families by primary target: ground transport and machinery (road vehicles, trains, industrial machinery, cyclists, and static vehicle displays), maritime vessels, aircraft (including seaplanes), and people.
Each family stresses EO and IR evidence differently: maritime events contain glare, wakes, and similar-looking vessels; ground transport and machinery events contain small or partly hidden targets against warm pavement; aircraft events add large changes in viewpoint and scale. The people events contain no identifiable individuals (Appendix~\ref{app:responsible_release}). These families define evaluation scenes and carry no deployment recommendation.

We verify EO--IR pairing at the event level. 
Annotators check that the same target-centric event appears in the overlapping EO and IR observations, remove clips with severe temporal or field-of-view mismatch, and treat residual differences in viewpoint, scale, sensor response, and overlap as part of the benchmark difficulty. 
As a result, RAVEN-Bench evaluates cross-spectral reasoning under realistic non-identical viewpoints, with no assumption of registered EO--IR pixels.

We retain clips that contain a target-centric event, event-level EO--IR correspondence, and sufficient visual evidence for at least one evaluation setting. 
We remove clips with missing streams, severe blur or instability, ambiguous event boundaries, insufficient target evidence, sensitive metadata, or personally identifiable information.
Retained clips are manually screened for identifiable people and readable identifiers; Appendix~\ref{app:responsible_release} gives the screening and redaction rules.
Section~\ref{sec:task_design} formalizes the evidence conditions used for evaluation.

\subsection{QA Construction and Quality Control}
\label{sec:qa_construction}

For each EO--IR pair, we construct twelve four-way multiple-choice QA items around the same target-centric event. 
This gives each event equal weight while probing target perception, temporal dynamics, modality selection, and event-level reasoning. 
Each item includes a ground-truth answer and diagnostic metadata used only for scoring, including modality requirement, capability level, reasoning tags, group family, and group step.

QA construction is human-authored, AI-assisted, and independently audited. 
Human annotators inspect the paired videos and write questions against the benchmark taxonomy. 
AI assistance is limited to wording refinement, option clarity, distractor review, and ambiguity flagging. 
It is not used to determine visual facts, answers, modality requirements, or capability/group labels.

Quality control filters out unsupported or ambiguous items, modality-label mismatches, language-prior shortcuts, and questions that depend on pixel-level registration, fine-grained geolocation, platform parameters, or unavailable sensor details. 
Grouped questions are kept only when they refer to the same target-centric event and support the diagnostics in Section~\ref{sec:task_design}. 
Screening categories, full guidelines, and metadata are provided in Appendix~\ref{app:annotation_protocol} and Appendix~\ref{app:metadata_schema}.

After these standard checks, a second, independent audit covers all 1,200 items: auditors answer each item without hidden annotations, then check the answer key, required sensor, evidence interval, distractors, shortcut risk, and group validity, with a third reviewer resolving disagreements.
The audit yields 96.5\% answer-key, 94.4\% modality, and 93.6\% group-validity agreement. All results use the audited annotations (Appendix~\ref{app:dataset_versions}).

\subsection{Dataset Statistics}
\label{sec:dataset_statistics}

\autoref{fig:datastatistics} summarizes the main dataset statistics. 
RAVEN-Bench contains 200 modality-specific videos organized into 100 EO--IR pairs.
The total modality-stream duration is 17,939.0 seconds (about 5 hours), with 9,027.2 seconds from EO and 8,911.8 seconds from IR, giving a mean stream duration of 89.7 seconds over the 200 single-modality streams.

By primary target, the 100 pairs comprise 59 ground transport and machinery, 27 maritime, 11 aircraft, and 3 people events.
Together, they carry 94 distinct fine-grained scenario labels across land, water, and airfield settings.
At the annotation level, the 1,200 four-way QA items are labeled by modality requirement, capability level, and reasoning tags for EO-only, IR-only, and paired EO+IR evaluation.

\section{Evaluation Task Design}
\label{sec:task_design}

RAVEN-Bench tests whether MLLMs can reason over paired aerial EO--IR videos in ways that are target-centric, temporal, and sensitive to the sensing modality. 
The annotation metadata defined in Section~\ref{sec:raven} specifies the evidence conditions, capability levels, and group structure used by the diagnostic protocol. 
Raw question-level accuracy is a weak signal in this setting because the decisive evidence may be small, brief, modality-specific, or distributed across the EO and IR streams.

\begin{figure*}[t]
    \centering
    \includegraphics[width=\textwidth]{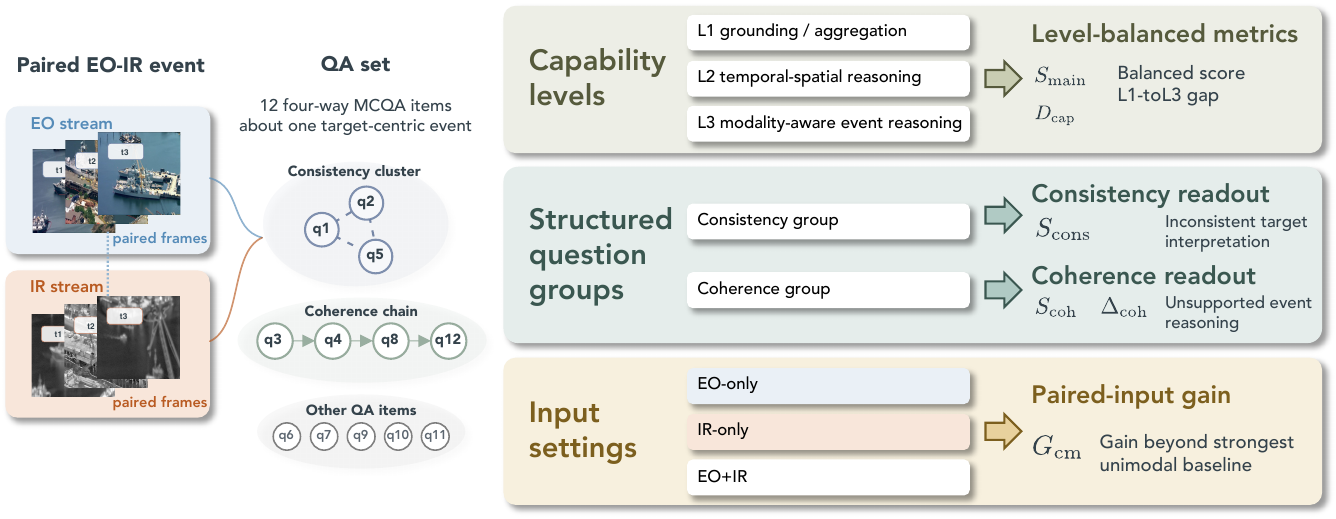}
\caption{
    Diagnostic evaluation design. RAVEN-Bench organizes a paired EO--IR event into a shared QA set and reads out three diagnostic signals: target consistency from consistency groups, evidence-chain coherence from coherence groups, and paired-input gain from EO-only, IR-only, and EO+IR settings. Capability levels and evidence-condition slices are used for balanced scoring and slice-level analysis.
    }
    \label{fig:diagnostic_design}
\end{figure*}

The evaluation focuses on three questions: whether the model keeps a stable interpretation of the same target, whether event-level answers are supported by earlier evidence, and whether paired EO+IR input improves over the stronger unimodal stream. Consistency groups, coherence groups, and modality-conditioned inputs provide these readouts, with capability levels and evidence-condition slices used for stratified analysis.
\autoref{fig:diagnostic_design} summarizes this target--probe--metric design.

\subsection{Task Formulation}
\label{sec:task_formulation}

Each RAVEN-Bench sample consists of a paired EO--IR video observation and twelve four-way multiple-choice QA items built around the same paired event. 
Given a visual input \(V\) and a question \(q_k\), a model predicts one answer option:
\begin{equation}
    \hat{y}_k = f(V, q_k), \quad \hat{y}_k \in \{A,B,C,D\}.
\end{equation}
The input \(V\) is provided in one of three modality-conditioned settings: the EO stream, the IR stream, or the paired EO+IR streams. 
All three settings use the same exported question set. 
Question-level modality requirements are used only after inference to compute diagnostic slices; they do not change the question text or the answer options shown to the model.

Models receive only the visual input, the natural-language question, and the answer options. 
Capability labels, modality requirements, group identifiers, reasoning skill tags, and other annotations are reserved for scoring and analysis, and are not provided during inference. 
RAVEN-Bench operates at the video-QA level, without pixel-level registration or sensor-fusion supervision. 
The EO and IR videos are paired observations of the same aerial event, and models are evaluated on whether they can identify relevant evidence, track target dynamics, and reason over modality-dependent observations.

\subsection{Diagnostic Structure}
\label{sec:diagnostic_structure}

We next describe how the annotations and input settings instantiate these diagnostics.

\paragraph{Structured question groups.}
\label{sec:group_structure}
Beyond individual QA items, RAVEN-Bench uses two group types tied to the first two diagnostic targets. 
A consistency group is unordered and contains related questions about the same target or event, so it tests whether a model gives compatible answers across related observations. 
A coherence group is ordered as a short reasoning chain: earlier questions verify intermediate evidence, and later questions ask for conclusions that depend on that evidence. 
The two group types use different scores because stable target interpretation and evidence-supported event reasoning expose different failure patterns. 
In the retained benchmark, every QA item belongs to either a consistency group or a coherence group. 
Level-wise and main-score metrics are still computed over individual QA items, and the scorer also supports ungrouped items for future extensions.

\paragraph{Modality-conditioned inputs.}
\label{sec:eo_ir_settings}
To test paired-input benefit, each model is evaluated under EO-only, IR-only, and EO+IR settings using the same question set. 
In the EO-only setting, the model receives only the electro-optical stream. 
In the IR-only setting, it receives only the infrared stream. 
In the EO+IR setting, it receives both paired streams with modality labels.
All settings use 16 synchronized event-relative timestamps per modality: EO-only and IR-only inputs contain 16 frames, and EO+IR inputs contain the matched EO and IR frames at each timestamp, up to 32 frames.
Input settings specify which streams are provided to the model. 
Evidence conditions describe which modality contains the decisive evidence for a question or evaluation slice. 
We use modality requirement as the item-level annotation field and evidence condition as the evaluation slice induced by that field. 
This design allows us to compare visible-spectrum evidence use, thermal evidence use, and measurable benefit from paired EO+IR input under matched questions. 
Sampling, prompt templates, input formatting, and parsing details are in Section~\ref{sec:experiments} and Appendix~\ref{app:model_details}.

\paragraph{Stratification metadata.}
\label{sec:capability_levels}
Capability levels, evidence conditions, and reasoning tags serve as a balancing and analysis layer. 
Level~1 covers grounding and cross-spectral evidence aggregation, Level~2 covers temporal/spatial reasoning under modality shift, and Level~3 covers modality-aware event reasoning. 
These levels define the macro-averaged main score and the capability gap. 
Evidence conditions and reasoning tags support slice-level analyses without changing model prompts.

\subsection{Metrics}
\label{sec:metrics}

All questions have four answer choices, which enables deterministic scoring. 
We use the diagnostics above for capability-balanced, group-level, and modality-conditioned analysis.

\paragraph{Main score.}
Let \(S_{L1}\), \(S_{L2}\), and \(S_{L3}\) denote the level-wise scores for the three capability levels. 
Each level score is computed as the average question accuracy over items assigned to that level. 
The main benchmark score is the macro-average:
\begin{equation}
S_{\mathrm{main}}=\frac{1}{3}(S_{L1}+S_{L2}+S_{L3}).
\end{equation}
This prevents a single level from dominating the aggregate score. 
We report it under EO-only, IR-only, and EO+IR inputs as \(S_{\mathrm{main}}^{\mathrm{EO}}\), \(S_{\mathrm{main}}^{\mathrm{IR}}\), and \(S_{\mathrm{main}}^{\mathrm{EO+IR}}\).

\paragraph{Consistency score.}
For a consistency group \(g\) with \(K_g\) questions and \(N_g\) correct predictions:
\begin{equation}
s_{\mathrm{cons}}(g)=\left(\frac{N_g}{K_g}\right)^2 .
\end{equation}
The squared form penalizes fragmented correctness while preserving partial credit. 
We report \(S_{\mathrm{cons}}\) as the macro-average over consistency groups.

\paragraph{Coherence score.}
For a coherence group \(g\), questions are ordered by reasoning step. 
We use coherence to mean prefix correctness over ordered evidence chains.
Let \(L_g\) be the number of consecutive correct answers from the first step before the first error. 
We define
\begin{equation}
s_{\mathrm{coh}}(g)=\frac{L_g}{K_g}.
\end{equation}
Later correct answers do not compensate for earlier failures because the final conclusion lacks required intermediate support. 
We report \(S_{\mathrm{coh}}\) as the macro-average over coherence groups.

\paragraph{Cross-modal gain.}
We measure paired EO+IR benefit beyond unimodal evidence as
\begin{equation}
G_{\mathrm{cm}}
=
S_{\mathrm{main}}^{\mathrm{EO+IR}}
-
\max\left(S_{\mathrm{main}}^{\mathrm{EO}}, S_{\mathrm{main}}^{\mathrm{IR}}\right).
\end{equation}
Positive \(G_{\mathrm{cm}}\) indicates paired-input improvement over the stronger unimodal condition. 
Near-zero or negative values are consistent with unimodal reliance, input-format limitations, or failure to use paired evidence. 
We report \(G_{\mathrm{cm}}\) globally and on EO--IR evidence-condition slices. 
Because \(G_{\mathrm{cm}}\) is an aggregate difference, it indicates paired-input gain and should not be read as proof of cross-modal integration.

\paragraph{Derived diagnostic gaps.}
We also report two diagnostic gaps. 
The capability gap
\begin{equation}
D_{\mathrm{cap}} = S_{L1}-S_{L3}
\end{equation}
measures degradation from local grounding to modality-aware event reasoning. 
The coherence gap
\begin{equation}
\Delta_{\mathrm{coh}} = S_{\mathrm{main}}^{\mathrm{EO+IR}} - S_{\mathrm{coh}}
\end{equation}
measures how much paired-input main performance overestimates ordered evidence-chain coherence. 
Unless otherwise specified, these diagnostics are computed under paired EO+IR input. 
Appendix~\ref{app:metric_details} gives additional aggregation details.

\paragraph{Uncertainty.}
We treat small point-estimate differences descriptively and avoid significance claims based only on isolated score gaps. 
Appendix~\ref{app:bootstrap_method} specifies the clustered bootstrap procedure used by the evaluation code. 
Question-level and level-wise metrics are resampled at the EO--IR video-pair level, while group metrics are resampled at the group level. 
This procedure avoids treating correlated QA items from the same paired observation as independent samples.

\section{Experiments}
\label{sec:experiments}

\subsection{Experimental Setup}
\label{sec:exp_setup}

We evaluate frontier API models, recent open-weight video-capable MLLMs, three remote-sensing VLMs (UniRS~\citep{li2024unirs}, SkyEyeGPT~\citep{zhan2024skyeyegpt}, and EarthDial~\citep{soni2025earthdial}), and non-visual controls in a zero-shot four-way multiple-choice setting. 
Open-weight models are grouped by total parameter scale. 
For MoE models, we use total parameters rather than activated parameters.

All models use the same EO-only, IR-only, and EO+IR settings, question set, option order, prompt template, and temporal sampling protocol (16 synchronized event-relative timestamps per modality). 
Each model is asked to return one option from \texttt{A}--\texttt{D}. 
We parse the first valid option deterministically and count invalid outputs as incorrect. 
The constrained evidence budget is used to isolate diagnostic behavior; serving efficiency and native-video utilization are outside the comparison. 
Appendix~\ref{app:experimental_protocol} gives prompts, sampling, formatting, parsing, and scoring details, while Appendix~\ref{app:robustness_controls} reports controls and human-reference results.

\subsection{Main Results}
\label{sec:main_results}

\autoref{tab:main_results} reports the main RAVEN-Bench results. 
The model groups are chosen to cover frontier API systems, open-weight checkpoints, hosted open-weight systems, and remote-sensing VLMs. 
We use these results to study paired-input gain, capability imbalance, and grouped consistency/coherence, rather than to produce a single leaderboard.

\newcommand{\metricup}[1]{\(#1\,{\scriptstyle\uparrow}\)}
\newcommand{\metricdown}[1]{\(#1\,{\scriptstyle\downarrow}\)}

\begin{table*}[t]
\centering
\caption{
Main diagnostic results. Higher is better except for $D_{\mathrm{cap}}$ and $\Delta_{\mathrm{coh}}$; \best{bold} and \second{underlined} mark best and second-best entries within this constrained evidence-budget protocol, not absolute model capability across serving interfaces.
}
\label{tab:main_results}
\setlength{\tabcolsep}{3.2pt}
\renewcommand{\arraystretch}{1.28}
\resizebox{\textwidth}{!}{%
\begin{tabular}{@{}lccccccccccc@{}}
\toprule
& \multicolumn{4}{c}{\textbf{Modality setting}} 
& \multicolumn{4}{c}{\textbf{Capability level under EO+IR}} 
& \multicolumn{3}{c}{\textbf{Consistency and coherence}} \\
\cmidrule(lr){2-5}
\cmidrule(lr){6-9}
\cmidrule(l){10-12}
\textbf{Model} 
& \metricup{S^{\mathrm{EO}}}
& \metricup{S^{\mathrm{IR}}}
& \metricup{S^{\mathrm{EO+IR}}}
& \metricup{G_{\mathrm{cm}}}
& \metricup{S_{L1}}
& \metricup{S_{L2}}
& \metricup{S_{L3}}
& \metricdown{D_{\mathrm{cap}}}
& \metricup{S_{\mathrm{cons}}}
& \metricup{S_{\mathrm{coh}}}
& \metricdown{\Delta_{\mathrm{coh}}} \\
\midrule

\groupheader{Frontier API models}
GPT-5.4~\citep{openai2026gpt54} 
& \second{45.2} & 41.6 & 46.1 & 0.9 
& \second{51.8} & 46.7 & 39.8 & 12.0 
& 30.6 & 22.8 & 23.3 \\
Gemini 3.1 Pro~\citep{google2026gemini31pro} 
& \best{46.2} & \best{44.9} & \best{49.6} & 3.4 
& \best{52.2} & \best{52.2} & \second{44.4} & 7.8 
& \second{36.8} & \second{30.2} & 19.4 \\
Gemini 3 Flash~\citep{google2025gemini3flash} 
& 40.7 & 43.3 & 42.6 & -0.7 
& 46.1 & 43.3 & 38.4 & 7.7 
& 25.4 & 20.3 & 22.3 \\

\midrule
\groupheader{Open-weight models, model size $\leq$4B}
SmolVLM2-2.2B-Instruct~\citep{marafioti2025smolvlm} 
& 30.4 & 33.1 & 31.7 & -1.4 
& 36.8 & 30.7 & 27.6 & 9.2 
& 13.2 & 8.3 & 23.4 \\
Qwen3-VL-2B-Instruct~\citep{bai2025qwen3vl} 
& 35.4 & 37.2 & 38.4 & 1.2 
& 41.9 & 39.8 & 33.5 & 8.4 
& 18.6 & 13.8 & 24.6 \\
VideoLLaMA3-2B~\citep{zhang2025videollama3} 
& 31.1 & 33.5 & 33.9 & 0.4 
& 35.8 & 34.3 & 31.6 & 4.2 
& 16.7 & 14.5 & 19.4 \\
Video-XL-Pro-3B~\citep{liu2025videoxlpro} 
& 32.6 & 34.8 & 35.7 & 0.9 
& 37.9 & 35.8 & 33.4 & 4.5 
& 20.3 & 17.1 & 18.6 \\

\midrule
\groupheader{Open-weight models, 4B $<$ model size $\leq$9B}
Qwen3.5-9B-Instruct~\citep{qwen2026qwen35} 
& 39.7 & 39.4 & 43.3 & 3.6 
& 45.3 & 46.4 & 38.2 & 7.1 
& 23.5 & 18.2 & 25.1 \\
Qwen3-VL-8B-Instruct~\citep{bai2025qwen3vl} 
& 41.9 & 38.7 & 43.3 & 1.4 
& 46.6 & 44.5 & 38.8 & 7.8 
& 24.7 & 19.7 & 23.6 \\
InternVL3.5-8B-Instruct~\citep{wang2025internvl35} 
& 38.8 & 40.5 & 39.9 & -0.6 
& 43.2 & 40.6 & 35.9 & 7.3 
& 20.7 & 16.2 & 23.7 \\
MiniCPM-V~4.5~\citep{yu2025minicpmv45} 
& 37.8 & 37.5 & 41.7 & \second{3.9} 
& 44.4 & 41.8 & 38.9 & 5.5 
& 24.5 & 19.5 & 22.2 \\
Molmo2-8B~\citep{ai22026molmo2} 
& 35.1 & 41.8 & 41.2 & -0.6 
& 47.9 & 40.1 & 35.6 & 12.3 
& 21.9 & 15.1 & 26.1 \\
Keye-VL-1.5-8B~\citep{yang2025keyevl15} 
& 38.6 & 38.9 & 43.4 & \best{4.5} 
& 46.1 & 43.3 & 40.8 & 5.3 
& 26.1 & 21.5 & 21.9 \\
MiMo-VL-7B-RL-2508~\citep{coreteam2025mimovl} 
& 40.2 & 40.8 & 41.9 & 1.1 
& 44.7 & 43.5 & 37.5 & 7.2 
& 23.8 & 17.8 & 24.1 \\
VideoLLaMA3-7B~\citep{zhang2025videollama3} 
& 39.7 & 37.8 & 42.4 & 2.7 
& 44.8 & 41.6 & 40.8 & 4.0 
& 25.2 & 21.9 & 20.5 \\
InternVideo2.5-Chat-8B~\citep{wang2025internvideo25} 
& 37.6 & 39.8 & 43.5 & 3.7 
& 45.0 & 43.6 & 41.9 & \second{3.1} 
& 27.6 & 23.8 & 19.7 \\
mPLUG-Owl3-7B~\citep{ye2024mplugowl3} 
& 37.1 & 36.6 & 37.7 & 0.6 
& 41.7 & 40.1 & 31.3 & 10.4 
& 17.9 & 12.5 & 25.2 \\
LLaVA-Video-7B-Qwen2~\citep{zhang2024llavavideo} 
& 29.3 & 27.6 & 29.8 & 0.5 
& 31.8 & 31.5 & 26.1 & 5.7 
& 8.4 & 6.9 & 22.9 \\

\midrule
\groupheader{Open-weight models, 9B $<$ model size $\leq$40B}
Kimi-VL-A3B-Instruct~\citep{kimi2025kimivl} 
& 41.7 & 40.2 & 42.1 & 0.4 
& 48.8 & 41.0 & 36.5 & 12.3 
& 23.2 & 15.5 & 26.6 \\
Qwen3-VL-30B-A3B-Instruct~\citep{bai2025qwen3vl} 
& 41.6 & 43.3 & 45.9 & 2.6 
& 48.5 & 47.9 & 41.3 & 7.2 
& 28.9 & 23.4 & 22.5 \\
InternVL3.5-30B-A3B~\citep{wang2025internvl35} 
& 40.4 & 42.9 & 43.8 & 0.9 
& 46.8 & 44.9 & 39.7 & 7.1 
& 27.3 & 22.1 & 21.7 \\

\midrule
\groupheader{Hosted open-weight models, model size $>$40B}
Qwen3-VL-235B-A22B-Instruct~\citep{bai2025qwen3vl} 
& 44.2 & 44.1 & \second{47.8} & 3.6 
& 50.0 & \second{50.3} & 43.1 & 6.9 
& 33.4 & 27.9 & 19.9 \\
InternVL3.5-241B-A28B~\citep{wang2025internvl35} 
& 43.0 & \second{44.8} & 45.5 & 0.7 
& 47.9 & 46.8 & 41.8 & 6.1 
& 32.7 & 27.1 & \second{18.4} \\

\midrule
\groupheader{Remote-sensing VLMs}
UniRS~\citep{li2024unirs} 
& 43.2 & 38.6 & 47.0 & 3.8 
& 46.2 & 50.2 & \best{44.6} & \best{1.6} 
& \best{37.2} & \best{33.4} & \best{13.6} \\
SkyEyeGPT~\citep{zhan2024skyeyegpt} 
& 38.4 & 34.2 & 39.1 & 0.7 
& 41.8 & 39.6 & 35.9 & 5.9 
& 24.1 & 19.8 & 19.3 \\
EarthDial~\citep{soni2025earthdial} 
& 40.5 & 37.9 & 42.3 & 1.8 
& 44.3 & 43.1 & 39.5 & 4.8 
& 28.4 & 23.6 & 18.7 \\
\bottomrule
\end{tabular}%
}
\end{table*}

\paragraph{Finding 1: aggregate performance overestimates grouped evidence reliability.}
Across the 26 evaluated models, \(S_{\mathrm{cons}}\) is lower than \(S_{\mathrm{main}}^{\mathrm{EO+IR}}\) by \(16.9\) points on average, and \(S_{\mathrm{coh}}\) is lower by \(21.8\) points. 
Every model in Table~\ref{tab:main_results} has a positive gap on both grouped diagnostics. 
Gemini~3.1~Pro~\citep{google2026gemini31pro}, for example, drops from \(49.6\%\) \(S_{\mathrm{main}}^{\mathrm{EO+IR}}\) to \(36.8\%\) \(S_{\mathrm{cons}}\) and \(30.2\%\) \(S_{\mathrm{coh}}\). 
UniRS~\citep{li2024unirs} has the strongest grouped scores even though it does not have the highest aggregate accuracy. 
These results indicate that grouped reliability captures failures that isolated question-level accuracy can hide.

\paragraph{Finding 2: paired EO+IR gains require controlled interpretation.} We interpret \(G_{\mathrm{cm}}\) together with input-perturbation controls, since a positive paired-input gain does not by itself establish sensor fusion.
\(G_{\mathrm{cm}}\) varies substantially across models: Keye-VL-1.5-8B~\citep{yang2025keyevl15}, MiniCPM-V~4.5~\citep{yu2025minicpmv45}, and UniRS~\citep{li2024unirs} show the largest positive point-estimate gains, while SmolVLM2-2.2B~\citep{marafioti2025smolvlm}, Molmo2-8B~\citep{ai22026molmo2}, and Gemini~3~Flash~\citep{google2025gemini3flash} show no positive gain. 
We treat positive values as directional paired-input improvements, with sensor-fusion claims left to the control analyses. 
In the control subset, aligned EO+IR reaches \(G_{\mathrm{ctrl}}=+3.1\) and \(S_{\mathrm{coh}}=23.4\). 
Duplicated EO+EO/IR+IR inputs yield smaller gains (\(+0.7,+0.3\)) and lower coherence (20.6, 21.1), while shuffled or temporally misaligned EO+IR reduces gain below the stronger unimodal baseline. 
These controls show that temporal and cross-spectral correspondence matter under this protocol.

\paragraph{Finding 3: local evidence performance is a weak proxy for event reasoning.}
The capability gap \(D_{\mathrm{cap}}=S_{L1}-S_{L3}\) measures how much performance drops from local grounding to modality-aware event reasoning. 
Kimi-VL-A3B~\citep{kimi2025kimivl} and Molmo2-8B~\citep{ai22026molmo2} have high Level~1 scores but \(D_{\mathrm{cap}}=12.3\). 
UniRS~\citep{li2024unirs} has the smallest gap (\(1.6\)) despite not having the highest aggregate EO+IR score. 

\subsection{Evidence-Condition Diagnostics}
\label{sec:diagnostic_analysis}

\begin{figure}[t]
    \centering
    \includegraphics[trim=1cm 1.4cm 1cm 1.3cm, clip, width=\linewidth]{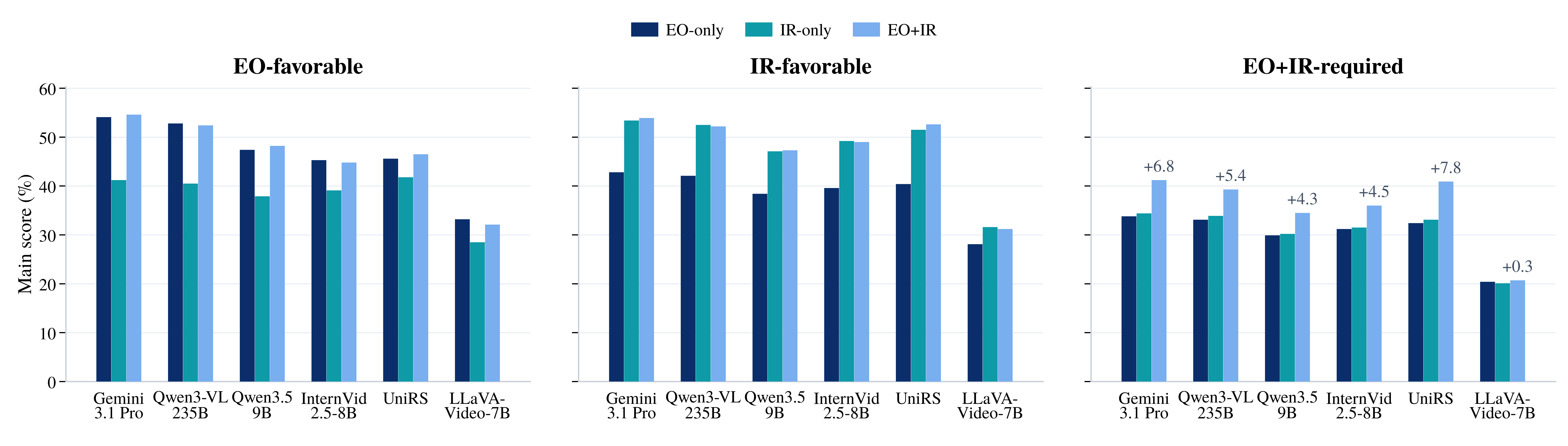}
    \caption{
    Evidence-condition slices for representative models. Bars compare EO-only, IR-only, and EO+IR inputs. Numbers above EO+IR bars in the EO+IR-required panel show \(G_{\mathrm{cm}}\) over the stronger unimodal input.
    }
    \label{fig:modality_slices}
\end{figure}

\autoref{fig:modality_slices} analyzes representative models across \texttt{EO-favorable}, \texttt{IR-favorable}, and \texttt{EO+IR-required} evidence conditions. 
This slice-level view tests whether models respond to the modality that contains the decisive evidence and whether paired-input gains appear when both streams are needed.

In the EO-favorable and IR-favorable slices, the annotated favorable modality separates the two unimodal inputs: EO-only exceeds IR-only on EO-favorable items, while IR-only exceeds EO-only on IR-favorable items. 
Paired EO+IR stays within about one point of the favorable unimodal input, since it includes that stream plus additional context. 
In the EO+IR-required slice, paired EO+IR exceeds both unimodal inputs for all representative models, giving the clearest slice-level evidence for paired-input gain. 
UniRS~\citep{li2024unirs} obtains the largest gain (\(G_{\mathrm{cm}}=7.8\)), followed by Gemini~3.1~Pro~\citep{google2026gemini31pro} (\(6.8\)) and Qwen3-VL-235B~\citep{bai2025qwen3vl} (\(5.4\)). 
LLaVA-Video-7B-Qwen2~\citep{zhang2024llavavideo} shows little benefit (\(0.3\)). 
These slices help distinguish unimodal skill, modality arbitration, and paired-input gain.

\subsection{Validity and Robustness Controls}
\label{sec:validity_robustness}

Appendix~\ref{app:robustness_controls} reports calibration, text-only, human-reference, input-perturbation, and frame-budget controls. 
The non-visual controls remain low: the best text-only setting reaches \(S_{\mathrm{main}}=28.4\) and \(S_{\mathrm{coh}}=9.8\), while the human-reference subset reaches 82.3 and 69.6 as an answerability calibration with retained EO--IR clips. 
The main paired-input control result follows the same pattern as the main evaluation. 
Aligned EO+IR gives the strongest control gain (\(G_{\mathrm{ctrl}}=+3.1\)); duplicated EO+EO/IR+IR inputs give much smaller gains (\(+0.7,+0.3\)); and shuffled or temporally misaligned EO+IR falls below the stronger unimodal baseline. 
Increasing the synchronized timestamp budget from \(T=16\) to \(T=32\) changes \(S_{\mathrm{main}}^{\mathrm{EO+IR}}\) only from 43.5 to 44.1 and \(S_{\mathrm{coh}}\) from 23.4 to 23.9. 
These controls support aligned EO--IR correspondence as the diagnostic condition while keeping claims about native-video utilization and sensor fusion bounded.
A free-form variant with human-judged answers and evidence statements keeps the same ordering (Main 34.8, Consistency 24.6, Coherence 19.3, Answer+Evidence 27.1; Appendix~\ref{app:free_form}), so the grouped gaps do not depend on the multiple-choice format.

\subsection{Observed Error Mechanisms}
\label{sec:failure_modes}

A manual audit identifies four recurring error mechanisms: aerial-platform ego-motion mistaken as target motion, tiny-target loss after resizing or sparse sampling, EO/IR distractors such as glare or hot backgrounds, and failed binding of complementary EO and IR evidence. 
The first two primarily disrupt temporal tracking and ordered coherence, while the latter two expose failures to select and bind the correct sensor evidence for the same target. 
Examples are in Appendix~\ref{app:failure_cases}.

\section{Conclusion}
\label{sec:conclusion}

RAVEN-Bench brings event-level EO--IR pairing, controlled input settings, consistency and coherence groups, and evidence-condition slices into a single video-QA evaluation.
The results show that aggregate accuracy can mask unstable target interpretation, unsupported event-level answers, and limited paired-input benefit beyond the stronger unimodal stream. 
We recommend reporting grouped reliability, capability gaps, and modality-conditioned slices alongside aggregate accuracy.
The gated release includes the evaluation code, Croissant metadata, and a dataset card; future versions will extend coverage while keeping the input conditions, correspondence controls, and grouped metrics fixed so that results remain comparable.

\paragraph{Limitations and responsible use.}
All videos come from one aircraft, one EO/IR gimbal, and one region in western North America, so the scores describe this sensing setup; condition subsets (weather, altitude, viewing geometry) remain too small for robustness comparisons.
Paired EO--IR aerial video is dual-use. RAVEN-Bench provides no spatial, trajectory, geolocation, or flight-path data; access is gated, and the data release terms limit use to research and non-commercial purposes and prohibit re-identification and the surveillance, tracking, or profiling of specific individuals. These safeguards cannot prevent every misuse after download; Appendix~\ref{app:responsible_release} details the remaining risk and the maintenance plan.

\begin{ack}
This work was supported by the Mitacs Accelerate Program (Grant No. IT42711), by TerraSense Analytics, and in part by grants from the US National Science Foundation (OAC-2118240, HDR Institute: Imageomics). We thank the University of British Columbia's Survive and Thrive Applied Research (UBC STAR) initiative for its support of this work. This work uses data provided by TerraSense Analytics Ltd. (\mbox{\url{www.terrasense.ca}}).
\end{ack}

{\small
\bibliographystyle{unsrtnat}
\bibliography{reference}
}


\newpage

\appendix

\section{Dataset Documentation and Annotation Protocol}
\label{app:dataset_annotation}

This appendix provides the dataset-level information needed to audit RAVEN-Bench and reproduce its diagnostic evaluation. The benchmark consists of paired EO--IR aerial video observations and four-way multiple-choice QA items. Metadata are used for dataset organization, quality control, and scoring; they are not provided to models during inference.

\subsection{Dataset Composition and Annotation-Derived Coverage}
\label{app:dataset_composition}

Table~\ref{tab:dataset_composition} summarizes the final benchmark composition, and Table~\ref{tab:retained_coverage} reports annotation-derived coverage over the retained evaluation set. RAVEN-Bench is intentionally compact and diagnostic rather than class-balanced at web scale: each retained EO--IR video pair contributes the same number of QA items, and the annotation metadata supports controlled slices by modality requirement, capability level, and group structure. The released annotation files document the retained benchmark; raw-pool rejection counts are therefore not reported because they require a separate screening log rather than the retained-set annotation package.

\begin{table}[t]
\centering
\small
\setlength{\tabcolsep}{5pt}
\caption{Dataset composition of RAVEN-Bench. Durations sum over the 200 single-modality streams.}
\label{tab:dataset_composition}
\begin{tabular}{p{0.42\linewidth}p{0.48\linewidth}}
\toprule
Statistic & Value \\
\midrule
EO--IR video pairs & 100 \\
Modality-specific video streams & 200 \\
QA items & 1,200 \\
Questions per video pair & 12 \\
Answer options per question & 4 \\
Total modality-stream duration & 17,939.0 s \\
EO stream duration & 9,027.2 s \\
IR stream duration & 8,911.8 s \\
Mean stream duration (200 streams) & 89.7 s \\
Scenario-family distribution & 59 ground transport and machinery / 27 maritime / 11 aircraft / 3 people \\
Fine-grained scenario labels (distinct \texttt{type} values) & 94 \\
Question-type tags & 20 tags / 1,963 assignments \\
\bottomrule
\end{tabular}
\end{table}

\begin{table*}[t]
\centering
\small
\setlength{\tabcolsep}{5pt}
\caption{Annotation-derived retained-set coverage. Counts are computed from the 100 JSON annotation files in the retained benchmark.}
\label{tab:retained_coverage}
\begin{tabular}{p{0.38\linewidth}p{0.54\linewidth}}
\toprule
Audit item & Annotation-derived count \\
\midrule
Retained annotation files & 100 files \\
QA items with four answer options and one answer key & 1,200 QA items \\
Capability-level coverage & L1: 386; L2: 560; L3: 254 \\
Modality-requirement coverage & EO: 656; IR: 229; EO+IR: 315 \\
Grouped-question metadata & 1,200 QA items include group identifiers, group family, and group step \\
Consistency groups & 100 groups / 800 QA items / 8 QA items per group \\
Coherence groups & 100 groups / 400 QA items / 4 QA items per group \\
Temporal evidence references & 1,200 QA items include at least one EO or IR time reference (971 with EO, 544 with IR, 315 with both) \\
\bottomrule
\end{tabular}
\end{table*}

\begin{table}[t]
\centering
\small
\setlength{\tabcolsep}{5pt}
\caption{Question-type tag distribution over the 1,200 QA items (20 tags, 1,963 assignments). Percentages use the 1,200 items as the denominator; tags are multi-label, so percentages need not sum to 100\%.}
\label{tab:question_type_tags}
\begin{tabular}{lcc}
\toprule
Question-type tag & Count & QA items (\%) \\
\midrule
\texttt{spatial\_reasoning} & 350 & 29.17 \\
\texttt{temporal\_reasoning} & 348 & 29.00 \\
\texttt{entity\_grounding} & 322 & 26.83 \\
\texttt{cross\_modal\_phenomenon\_explanation} & 187 & 15.58 \\
\texttt{camera\_motion\_vs\_target\_motion} & 119 & 9.92 \\
\texttt{thermal\_evidence\_interpretation} & 115 & 9.58 \\
\texttt{evidence\_sufficiency} & 91 & 7.58 \\
\texttt{cross\_modal\_correspondence} & 81 & 6.75 \\
\texttt{ir\_contrast\_grounding} & 76 & 6.33 \\
\texttt{spatial\_temporal\_reasoning} & 75 & 6.25 \\
\texttt{group\_verdict} & 60 & 5.00 \\
\texttt{trajectory\_reasoning} & 40 & 3.33 \\
\texttt{temporal\_event\_state\_account} & 38 & 3.17 \\
\texttt{scene\_relation} & 31 & 2.58 \\
\texttt{identity\_continuity} & 10 & 0.83 \\
\texttt{distractor\_rejection} & 7 & 0.58 \\
\texttt{scene\_grounding} & 5 & 0.42 \\
\texttt{causal\_reasoning} & 4 & 0.33 \\
\texttt{evidence\_retrieval} & 3 & 0.25 \\
\texttt{occlusion\_reasoning} & 1 & 0.08 \\
\bottomrule
\end{tabular}
\end{table}

\subsection{Metadata Schema}
\label{app:metadata_schema}

RAVEN-Bench uses lightweight metadata for auditing and diagnostic scoring. Table~\ref{tab:video_metadata_schema} lists sample-level annotation fields, and Table~\ref{tab:question_metadata_schema} lists QA-level fields. Fields that expose event summaries, answer keys, option roles, rationales, or time references are used for auditing and scoring only; they are not provided to models during inference.

\begin{table}[t]
\centering
\small
\setlength{\tabcolsep}{4pt}
\caption{Sample-level annotation schema. These fields organize the retained EO--IR video pairs and are not model input.}
\label{tab:video_metadata_schema}
\begin{tabular}{p{0.24\linewidth}p{0.18\linewidth}p{0.50\linewidth}}
\toprule
Field & Type & Description \\
\midrule
\texttt{sample\_id} & string & Unique identifier for the EO--IR video pair. \\
\texttt{type} & categorical & Released fine-grained retained scenario label used for coverage analysis. \\
\texttt{main\_event} & text & Internal event descriptor used during annotation and auditing. \\
\texttt{event\_description} & text & Internal summary of the target-centric event. \\
\texttt{scenario\_type} & categorical & Legacy internal annotation descriptor used during QA construction; not used as an evidence-condition label or scoring slice. \\
\texttt{notes} & object & Annotation warnings, design notes, and acceptance gates for prompt-leakage and ambiguity checks. \\
\texttt{qa} & list & The twelve QA annotations associated with the video pair. \\
\bottomrule
\end{tabular}
\end{table}

The field \texttt{type} is the released fine-grained scenario label used for retained-set coverage analysis. The field \texttt{scenario\_type} is retained only as an internal annotation descriptor and should not be confused with item-level \texttt{modality\_requirement}, which defines the evidence-condition slices used for scoring.

\begin{table}[t]
\centering
\small
\setlength{\tabcolsep}{4pt}
\caption{Question-level metadata schema. These fields support scoring, slicing, and annotation auditing; they are not provided to models during inference.}
\label{tab:question_metadata_schema}
\begin{tabular}{p{0.24\linewidth}p{0.18\linewidth}p{0.50\linewidth}}
\toprule
Field & Type & Description \\
\midrule
\texttt{uid} & string & Unique identifier for the QA item within a sample. \\
\texttt{group\_id} & string & Identifier for the consistency or coherence group. \\
\texttt{group\_family} & categorical & \texttt{consistency} or \texttt{coherence}. \\
\texttt{group\_focus} & text & Short internal descriptor of the group target or evidence pattern. \\
\texttt{group\_step} & integer & Ordered step index used for grouped scoring. \\
\texttt{capability\_level} & categorical & \texttt{L1}, \texttt{L2}, or \texttt{L3}. \\
\texttt{question} & text & Natural-language question. \\
\texttt{options} & object & Four answer options labeled \texttt{A}--\texttt{D}. \\
\texttt{answer} & categorical & Ground-truth answer option. \\
\texttt{option\_roles} & object & Internal record of correct and distractor roles. \\
\texttt{question\_type} & list & Multi-label reasoning tags used for dataset analysis. \\
\texttt{time\_reference\_eo} & text & Relevant EO temporal interval, when applicable. \\
\texttt{time\_reference\_ir} & text & Relevant IR temporal interval, when applicable. \\
\texttt{modality\_requirement} & categorical & Item-level modality requirement, i.e., the required evidence modality: \texttt{EO}, \texttt{IR}, or \texttt{EO+IR}. \\
\bottomrule
\end{tabular}
\end{table}

\subsection{Annotation Protocol and Independent Audit}
\label{app:annotation_protocol}

The annotation workflow is human-authored, AI-assisted, and independently audited. Human annotators inspect paired EO and IR videos, identify visual evidence, and write candidate questions and answer options. AI assistance is restricted to wording refinement, option clarity, distractor review, and ambiguity flagging; it is not used to determine visual facts, correct answers, modality requirements, capability labels, or group assignments. All accepted items are checked against the EO--IR video evidence before inclusion.

Annotation labor is organized by role rather than crowdsourced at scale. Retained QA construction was completed by three trained annotators with computer-vision or remote-sensing familiarity; before authoring, they reviewed the benchmark taxonomy and calibrated on pilot EO--IR clips covering EO-only, IR-only, and EO+IR-required evidence. The independent audit covers all items and is conducted by reviewers who did not author the audited items (Appendix~\ref{app:dataset_versions}). The human-reference evaluation in Appendix~\ref{app:validity_controls} uses a stratified 120-item subset, with separate instructions and scoring: audit reviewers check answer keys, modality requirements, and group validity, whereas human-reference annotators answer model-facing questions using retained EO--IR clips without hidden metadata. Annotation and audit labor was performed by authors or institutionally supervised research staff as part of paid research roles; no unpaid crowdsourcing was used.

Annotators follow five rules: each item must be visually answerable from the provided evidence; each item must have exactly one correct option; distractors must be plausible but visually incorrect; the relevant temporal evidence must remain identifiable despite platform motion and target-scale variation; and items must not require exact pixel-level registration, fine-grained geolocation, platform parameters, or unavailable sensor details.

Table~\ref{tab:screening_categories} summarizes the screening categories used during annotation review. The retained annotation package records accepted items and required metadata, but the screening log was not maintained as a countable rejection ledger; we therefore report categories rather than rejection counts.

\begin{table}[t]
\centering
\small
\setlength{\tabcolsep}{4pt}
\caption{Annotation screening categories. These categories were used to reject or revise candidate QA items before inclusion; counts are not reported because the retained release package is not a rejection log.}
\label{tab:screening_categories}
\begin{tabular}{p{0.33\linewidth}p{0.56\linewidth}}
\toprule
Screening category & Rejection or revision rationale \\
\midrule
Unsupported visual evidence & The proposed answer cannot be verified from the supplied EO or IR video evidence. \\
Multiple plausible answers & More than one option remains visually defensible after reviewing the relevant time interval. \\
Modality requirement mismatch & The annotated required modality does not match where the decisive evidence appears. \\
Language-only shortcut & The answer can be inferred from wording, option priors, or commonsense without visual evidence. \\
Pixel-registration requirement & The item requires exact EO--IR pixel alignment rather than event-level correspondence. \\
Unavailable sensor or platform metadata & The item depends on geolocation, sensor parameters, flight path, altitude, or platform details not released for evaluation. \\
Sensitive metadata risk & The item or rationale exposes operational, identifying, or unnecessary collection details. \\
\bottomrule
\end{tabular}
\end{table}

As summarized in Table~\ref{tab:retained_coverage}, every QA item includes a capability level, modality requirement, group identifier, group family, group step, answer key, four answer options, question-type tags, and at least one EO or IR time reference. The notes field in each annotation file also records prompt-leakage safeguards: event descriptions, answer keys, option roles, rationales, and time-reference metadata are used for scoring and auditing but are not exported to model prompts.

\subsection{Full Independent Audit and Capability Assignment}
\label{app:dataset_versions}

\paragraph{Full audit.}
Every one of the 1,200 items is independently audited after the standard checks in Section~\ref{sec:qa_construction}. A reviewer first answers each item and locates its evidence without hidden annotations, then checks the answer key, required sensor, evidence interval, distractors, shortcut risk, and group validity. A third reviewer resolves disagreements. Table~\ref{tab:full_audit} reports the results.

\begin{table}[t]
\centering
\small
\setlength{\tabcolsep}{6pt}
\caption{Full independent audit of the 100 EO--IR pairs and 1,200 QA items.}
\label{tab:full_audit}
\begin{tabular}{lc}
\toprule
Audit result & Value \\
\midrule
Coverage & 100/100 pairs; 1,200/1,200 items \\
Answer-key agreement & 96.5\% \\
Modality agreement & 94.4\% \\
Group-validity agreement & 93.6\% \\
Wording or distractor revisions & 114 items (9.5\%) \\
Answer-key corrections & 42 items (3.5\%) \\
Modality corrections & 67 items (5.6\%) \\
Evidence-interval corrections & 53 items (4.4\%) \\
Group corrections & 77 items (6.4\%) \\
Removed or replaced items & 19 items (1.6\%) \\
\bottomrule
\end{tabular}
\end{table}

\paragraph{Capability assignment.}
Human annotators assign every item to L1, L2, or L3 with a single item-level rubric, without per-level quotas. The resulting distribution is 386 L1, 560 L2, and 254 L3 items (Table~\ref{tab:retained_coverage}).

\paragraph{Construction procedure.}
All pairs share the same question structure, annotation format, review procedure, metrics, and scoring code, and were constructed by the annotator team described in Appendix~\ref{app:annotation_protocol}. Event selection prioritized target categories and capture conditions that would otherwise be rare. Independent evidence review is the main time cost. A release of thousands of events would require a larger annotation team and tools for routine annotation checks, and human review of the visual evidence would remain part of the process.

\subsection{Representative QA Examples}
\label{app:qa_examples}

Table~\ref{tab:qa_examples} gives representative examples from released or anonymized samples. The examples are included to clarify the intended distinction between EO-only, IR-only, EO+IR-required, consistency-group, and coherence-group questions.
The question texts shown in the table are retained model-facing question texts rather than hidden-metadata summaries. Table columns such as type, level, evidence interval, and group label are explanatory appendix metadata only; models receive only the visual input, the question, and the four answer options. Wording such as ``chain'' or ``unsafe'' is used to ask for visually grounded temporal comparison or to avoid overclaiming an event interpretation, and does not expose capability labels, modality requirements, group membership, answer keys, option roles, or rationales.

\begin{table*}[t]
\centering
\small
\setlength{\tabcolsep}{4pt}
\caption{Representative QA examples from the retained annotations. Examples are drawn from \texttt{TerraSense\_3.json}.}
\label{tab:qa_examples}
\begin{tabular}{p{0.14\linewidth}p{0.12\linewidth}p{0.40\linewidth}p{0.26\linewidth}}
\toprule
Type & Level & Question or group item & Correct option / evidence \\
\midrule
EO-only & \texttt{L1} & In the first close EO pass, which ground relation places the aircraft? & A; EO 00:17--00:29. \\
IR-only & \texttt{L1} & In the close IR pass, where is the bright pavement relative to the aircraft body? & A; IR 00:28--00:52. \\
EO+IR-required & \texttt{L3} & Which cross-modal chain avoids the paint-color shortcut? & A; EO 00:17--02:12 and IR 00:28--01:25. \\
Consistency group & \texttt{L1/L2} & \texttt{TerraSense\_3} \texttt{C1}: eight grouped questions about orbit-induced screen-side changes and EO--IR placement cues. & Annotation answer keys for steps 1--8; EO and IR intervals vary by item. \\
Coherence group & \texttt{L2/L3} & \texttt{TerraSense\_3} \texttt{H1}: four ordered questions from unsafe EO motion interpretation to cross-modal full-clip reading. & Annotation answer keys for steps 1--4; EO 00:17--02:12 and IR 00:28--01:25. \\
\bottomrule
\end{tabular}
\end{table*}

\subsection{Group-Level Annotation Examples}
\label{app:group_examples}

Table~\ref{tab:consistency_group_example} gives an aircraft consistency group, and Table~\ref{tab:coherence_group_example_app} gives a maritime coherence group. 
The consistency group keeps the same target fixed while varying EO evidence, IR evidence, temporal scale changes, and cross-modal cue transfer. 
The coherence group is ordered: early steps verify supporting EO and IR evidence, and later steps ask for cross-modal and full-clip conclusions.
These examples are used for compact illustration; analogous consistency and coherence groups are included for maritime, ground-vehicle, and aircraft samples in the released annotations.

\begin{table*}[t]
\centering
\scriptsize
\setlength{\tabcolsep}{2.5pt}
\caption{Complete consistency group example from \texttt{TerraSense\_3.json}. The eight items all refer to the same aircraft-centered event and test whether a model preserves a stable target interpretation across local EO and IR cues, temporal changes, and cross-modal checks.}
\label{tab:consistency_group_example}
\begin{tabular}{cp{0.09\linewidth}p{0.07\linewidth}p{0.44\linewidth}p{0.22\linewidth}}
\toprule
Step & Req. & Level & Question & Consistency role \\
\midrule
1 & EO & L1 & In the first close EO pass, which ground relation places the aircraft? & Grounds the target in EO using nearby layout. \\
2 & IR & L1 & In the close IR pass, where is the bright pavement relative to the aircraft body? & Checks matching target placement in IR. \\
3 & EO & L1 & In the later EO pass after the orbit, which screen-side layout is visible? & Rechecks placement after viewpoint change. \\
4 & EO & L2 & Which EO sequence describes the pavement shifting around a grass-side aircraft? & Separates camera orbit from target travel. \\
5 & IR & L2 & Which IR sequence tracks the aircraft through the scale change? & Tests IR persistence of the same target. \\
6 & EO+IR & L2 & Which EO detail becomes a weak direct cue in IR? & Tests whether an EO cue transfers poorly to IR. \\
7 & EO+IR & L3 & Which chain keeps the aircraft off pavement in both modalities? & Requires a stable cross-modal target interpretation. \\
8 & EO+IR & L3 & Which cue argues against a taxi-along-pavement interpretation? & Tests the group-level event interpretation. \\
\bottomrule
\end{tabular}
\end{table*}

\begin{table*}[t]
\centering
\scriptsize
\setlength{\tabcolsep}{2.5pt}
\caption{Complete coherence group example from the maritime sample \texttt{TerraSense\_46.json}. The questions are scored as an ordered evidence chain: failures in early EO or IR evidence steps limit the prefix coherence credit for later event-level answers.}
\label{tab:coherence_group_example_app}
\begin{tabular}{cp{0.09\linewidth}p{0.07\linewidth}p{0.44\linewidth}p{0.22\linewidth}}
\toprule
Step & Req. & Level & Question & Coherence role \\
\midrule
1 & EO & L2 & When the view changes, which cue is checked against the nearby alternative? & Verifies EO tug-to-barge tow geometry against the cooler barge-pair alternative. \\
2 & IR & L2 & Which IR cue pairing appears in this interval? & Verifies warm hull or disturbed-water contrast paired with the tow geometry. \\
3 & EO+IR & L3 & Which combined EO/IR sequence account is visible? & Combines EO tow geometry with IR thermal/wake evidence. \\
4 & EO+IR & L3 & For the full clip, which cue chain is visible? & Tests the final maritime event-level conclusion over the full paired observation. \\
\bottomrule
\end{tabular}
\end{table*}

\section{Metric and Scoring Implementation}
\label{app:metric_details}

This appendix gives implementation details for the metrics defined in Section~\ref{sec:task_design}. All metrics are computed from parsed multiple-choice predictions. Unless otherwise specified, grouped metrics and derived gaps are computed under paired EO+IR input.

\subsection{Metric Definitions}
\label{app:metric_definitions}

Question-level accuracy is the fraction of QA items whose parsed prediction matches the ground-truth option. Level-wise scores are computed by averaging question accuracy within each capability level, and the main score is the macro-average
\begin{equation}
S_{\mathrm{main}} = \frac{1}{3}\left(S_{L1}+S_{L2}+S_{L3}\right).
\end{equation}
For a consistency group \(g\) with \(K_g\) questions and \(N_g\) correct predictions, we compute
\begin{equation}
s_{\mathrm{cons}}(g)=\left(\frac{N_g}{K_g}\right)^2,
\end{equation}
and report \(S_{\mathrm{cons}}\) as the macro-average over consistency groups. For an ordered coherence group, let \(L_g\) be the number of consecutive correct answers from the first step before the first error. We compute
\begin{equation}
s_{\mathrm{coh}}(g)=\frac{L_g}{K_g},
\end{equation}
and report \(S_{\mathrm{coh}}\) as the macro-average over coherence groups. Cross-modal gain is computed as
\begin{equation}
G_{\mathrm{cm}}=S_{\mathrm{main}}^{\mathrm{EO+IR}}-\max\left(S_{\mathrm{main}}^{\mathrm{EO}},S_{\mathrm{main}}^{\mathrm{IR}}\right).
\end{equation}

\subsection{Clustered Bootstrap Procedure}
\label{app:bootstrap_method}

The evaluation code supports nonparametric bootstrap resampling that preserves the clustered structure of the benchmark. 
For \(S_{\mathrm{main}}^{\mathrm{EO}}\), \(S_{\mathrm{main}}^{\mathrm{IR}}\), \(S_{\mathrm{main}}^{\mathrm{EO+IR}}\), level-wise scores, \(G_{\mathrm{cm}}\), \(D_{\mathrm{cap}}\), and \(\Delta_{\mathrm{coh}}\), one bootstrap draw samples the 100 EO--IR video pairs with replacement and includes all twelve QA items associated with each selected pair. 
For \(S_{\mathrm{cons}}\), \(S_{\mathrm{coh}}\), and consistency-sensitivity readouts, we resample groups with replacement within the corresponding group family. 
In the retained benchmark, each EO--IR pair contributes one consistency group and one coherence group, so group-level resampling also preserves the event-level clustering used by the grouped diagnostics. 
Derived quantities such as \(G_{\mathrm{cm}}\), \(D_{\mathrm{cap}}\), and \(\Delta_{\mathrm{coh}}\) are recomputed inside each bootstrap draw rather than formed by subtracting marginal interval endpoints. 
When intervals are computed, the default setting uses 10{,}000 bootstrap draws with a fixed random seed, with the 2.5th and 97.5th percentiles as interval endpoints. 
This procedure treats questions from the same paired observation or the same group as correlated rather than independent.

\subsection{Alternative Group-Level Consistency Readouts}
\label{app:consistency_sensitivity}

The squared consistency score in Section~\ref{sec:metrics} is intended to penalize fragmented correctness while retaining partial credit. 
For robustness auditing, the scorer also defines two alternative group-level readouts:
\begin{align}
s_{\mathrm{cons\mbox{-}linear}}(g) &= \frac{N_g}{K_g}, \\
s_{\mathrm{cons\mbox{-}square}}(g) &= \left(\frac{N_g}{K_g}\right)^2, \\
s_{\mathrm{cons\mbox{-}all}}(g) &= \mathbf{1}[N_g=K_g].
\end{align}
\noindent
The linear readout measures mean within-group correctness, the squared readout emphasizes fragmented correctness, and the all-correct readout measures strict group completion.
The main paper reports the squared readout as the primary consistency metric, and the released scorer can compute the linear and all-correct variants from the same group-level predictions. 
Table~\ref{tab:consistency_sensitivity} reports this sensitivity check on representative models using scores computed from the retained group-level predictions.

\begin{table}[t]
\centering
\small
\setlength{\tabcolsep}{5pt}
\caption{Consistency-score sensitivity for representative models. Scores are percentages and are computed from the retained group-level predictions.}
\label{tab:consistency_sensitivity}
\begin{tabular}{lccc}
\toprule
Model & $S_{\mathrm{cons\mbox{-}linear}}$ & $S_{\mathrm{cons\mbox{-}square}}$ & $S_{\mathrm{cons\mbox{-}all}}$ \\
\midrule
Gemini 3.1 Pro & 57.2 & 36.8 & 21.2 \\
Qwen3-VL-235B & 54.5 & 33.4 & 19.1 \\
UniRS & 57.9 & 37.2 & 23.4 \\
LLaVA-Video-7B-Qwen2 & 25.3 & 8.4 & 2.2 \\
\bottomrule
\end{tabular}
\end{table}

The same broad diagnostic ordering holds under a lenient linear readout, the primary squared readout, and a strict all-correct readout. This supports the conclusion that the grouped-reasoning contrast is not an artifact of the squared penalty alone.

\subsection{Prefix Coherence Example}
\label{app:coherence_example}

Prefix coherence assumes that an ordered group represents an evidence chain. 
Later correct answers therefore do not fully compensate for earlier failures, because the final answer may be correct for the wrong reason or by exploiting a shortcut rather than by preserving the required visual evidence. 
Table~\ref{tab:coherence_chain_example} shows a retained coherence group from \texttt{TerraSense\_3.json}. 
The first two steps verify EO and IR evidence needed to avoid confusing camera orbit with aircraft motion; the final steps ask for cross-modal and full-clip conclusions built on that evidence.

\begin{table*}[t]
\centering
\small
\setlength{\tabcolsep}{4pt}
\caption{Example coherence chain from a retained sample. If a model fails step 1 or 2, a correct step-4 answer is not counted as full evidence-supported coherence because the intermediate EO or IR evidence needed by the final conclusion was not established.}
\label{tab:coherence_chain_example}
\begin{tabular}{cp{0.14\linewidth}p{0.12\linewidth}p{0.42\linewidth}p{0.18\linewidth}}
\toprule
Step & Requirement & Level & Question & Evidence role \\
\midrule
1 & EO & L2 & Which early-to-late EO chain would be unsafe to treat as aircraft travel? & Distinguishes platform orbit from target travel in EO. \\
2 & IR & L2 & In the later IR view, which paved-surface layout surrounds the aircraft? & Checks the IR spatial layout around the same target. \\
3 & EO+IR & L3 & Which cross-modal chain avoids the paint-color shortcut? & Requires combining EO temporal evidence with IR evidence. \\
4 & EO+IR & L3 & Which full-clip reading uses the orbit and IR evidence together? & Tests the final event-level interpretation over the full paired observation. \\
\bottomrule
\end{tabular}
\end{table*}

\subsection{Group Handling Rules}
\label{app:group_rules}

Groups are scored only when they contain at least two QA items with the same \texttt{group\_id}. Consistency groups are unordered and use all questions assigned to the group. Coherence groups require valid \texttt{group\_step} indices; if a step index is missing or duplicated, the group is excluded from \(S_{\mathrm{coh}}\) and flagged for annotation review. In the retained benchmark, all QA items are assigned to either a consistency or coherence group; level-wise and main-score metrics are still computed over individual QA items. Future ungrouped items would remain included in question-level, level-wise, and main-score evaluation but would not contribute to \(S_{\mathrm{cons}}\) or \(S_{\mathrm{coh}}\).

\section{Experimental Protocol}
\label{app:experimental_protocol}

This appendix documents the inference protocol used for all model evaluations. Models receive only the visual input, the question, and four answer options. Capability labels, modality requirements, group identifiers, skill tags, and ground-truth answers are used only for scoring and analysis.

\subsection{Shared Inference Protocol}
\label{app:model_details}

Table~\ref{tab:model_details} summarizes the protocol-level settings shared across evaluated models. Tables~\ref{tab:model_serving_details} and~\ref{tab:model_serving_details_cont} summarize the serving records used for the evaluated systems, including API snapshots or checkpoint identifiers, interfaces, decoding settings, and hardware/runtime notes.

\begin{table}[!htbp]
\centering
\small
\setlength{\tabcolsep}{5pt}
\caption{Protocol-level inference settings shared across evaluations.}
\label{tab:model_details}
\begin{tabular}{p{0.25\linewidth}p{0.65\linewidth}}
\toprule
Setting & Protocol \\
\midrule
Task & Zero-shot four-way multiple-choice QA. \\
Input settings & EO-only, IR-only, and paired EO+IR. \\
Question set & The same retained 1,200 QA items and answer-option order are used across input settings. \\
Temporal budget & 16 synchronized event-relative timestamps per modality: 16 frames for EO-only and IR-only; the matched EO and IR frames at each timestamp, up to 32 modality-labeled frames, for EO+IR. \\
Visual annotations & No target boxes, crops, masks, capability labels, modality requirements, group identifiers, or answers are shown to models. \\
Prompt template & The input-specific templates in Appendix~\ref{app:prompt_templates}. \\
Output format & One answer option from \texttt{A}, \texttt{B}, \texttt{C}, and \texttt{D}. \\
Prediction parsing & First valid option extracted deterministically; invalid outputs are counted incorrect. \\
Judging & No external LLM judge is used. \\
\bottomrule
\end{tabular}
\end{table}

\begin{table}[!htbp]
\centering
\tiny
\setlength{\tabcolsep}{2.2pt}
\renewcommand{\arraystretch}{0.88}
\caption{Model-specific serving records, part 1.}
\label{tab:model_serving_details}
\resizebox{\textwidth}{!}{%
\begin{tabular}{p{0.16\linewidth}p{0.29\linewidth}p{0.18\linewidth}p{0.18\linewidth}p{0.22\linewidth}}
\toprule
Model & Checkpoint or API snapshot & Interface & Decoding & Hardware/runtime notes \\
\midrule
GPT-5.4 & OpenAI API snapshot \texttt{gpt-5.4-2026-03-05} & Responses API \texttt{/v1/responses}; sampled image frames & Temperature 0; max output 16 & Provider-hosted; hardware not disclosed. \\
Gemini 3.1 Pro & \texttt{gemini-3.1-pro-preview} & Gemini API; multimodal input & Temperature 0; max output 16 & Provider-hosted; hardware not disclosed. \\
Gemini 3 Flash & \texttt{gemini-3-flash-preview} & Gemini API; multimodal input & Temperature 0; max output 16 & Provider-hosted; hardware not disclosed. \\
SmolVLM2-2.2B & \texttt{HuggingFaceTB/SmolVLM2-2.2B-Instruct} & Transformers checkpoint; sampled frames & Greedy decoding; max new tokens 16 & Single NVIDIA A100 GPU; BF16 inference; single-GPU runtime. \\
Qwen3-VL-2B & \texttt{Qwen/Qwen3-VL-2B-Instruct} & Transformers checkpoint; sampled frames & Greedy decoding; max new tokens 16 & Single NVIDIA A100 GPU; BF16 inference; single-GPU runtime. \\
VideoLLaMA3-2B & \texttt{DAMO-NLP-SG/VideoLLaMA3-2B} & Transformers checkpoint; sampled frames & Greedy decoding; max new tokens 16 & Single NVIDIA A100 GPU; BF16 inference; single-GPU runtime. \\
Video-XL-Pro-3B & \texttt{MINT-SJTU/Video-XL-Pro-3B} & Official checkpoint; sampled frames & Greedy decoding; max new tokens 16 & Single NVIDIA A100 GPU; BF16 inference; single-GPU runtime. \\
Qwen3.5-9B & \texttt{Qwen/Qwen3.5-9B} & Transformers/vLLM checkpoint; sampled frames & Greedy decoding; max new tokens 16 & Single NVIDIA A100 GPU; BF16 inference; single-GPU runtime. \\
Qwen3-VL-8B & \texttt{Qwen/Qwen3-VL-8B-Instruct} & Transformers checkpoint; sampled frames & Greedy decoding; max new tokens 16 & Single NVIDIA A100 GPU; BF16 inference; single-GPU runtime. \\
InternVL3.5-8B & \texttt{OpenGVLab/InternVL3\_5-8B} & Transformers checkpoint; sampled frames & Greedy decoding; max new tokens 16 & Single NVIDIA A100 GPU; BF16 inference; single-GPU runtime. \\
MiniCPM-V 4.5 & \texttt{openbmb/MiniCPM-V-4\_5} & Transformers checkpoint; sampled frames & Greedy decoding; max new tokens 16 & Single NVIDIA A100 GPU; BF16 inference; single-GPU runtime. \\
Molmo2-8B & \texttt{allenai/Molmo2-8B} & Transformers checkpoint; sampled frames & Greedy decoding; max new tokens 16 & Single NVIDIA A100 GPU; BF16 inference; single-GPU runtime. \\
\bottomrule
\end{tabular}%
}
\end{table}

\begin{table}[!htbp]
\centering
\tiny
\setlength{\tabcolsep}{2.2pt}
\renewcommand{\arraystretch}{0.88}
\caption{Model-specific serving records, continued.}
\label{tab:model_serving_details_cont}
\resizebox{\textwidth}{!}{%
\begin{tabular}{p{0.16\linewidth}p{0.29\linewidth}p{0.18\linewidth}p{0.18\linewidth}p{0.22\linewidth}}
\toprule
Model & Checkpoint or API snapshot & Interface & Decoding & Hardware/runtime notes \\
\midrule
Keye-VL-1.5-8B & \texttt{Kwai-Keye/Keye-VL-1.5-8B} & Transformers checkpoint; sampled frames & Greedy decoding; max new tokens 16 & Single NVIDIA A100 GPU; BF16 inference; single-GPU runtime. \\
MiMo-VL-7B-RL-2508 & \texttt{XiaomiMiMo/MiMo-VL-7B-RL-2508} & Transformers checkpoint; sampled frames & Greedy decoding; max new tokens 16 & Single NVIDIA A100 GPU; BF16 inference; single-GPU runtime. \\
VideoLLaMA3-7B & \texttt{DAMO-NLP-SG/VideoLLaMA3-7B} & Transformers checkpoint; sampled frames & Greedy decoding; max new tokens 16 & Single NVIDIA A100 GPU; BF16 inference; single-GPU runtime. \\
InternVideo2.5-Chat-8B & \texttt{OpenGVLab/InternVideo2\_5\_Chat\_8B} & Transformers checkpoint; sampled frames & Greedy decoding; max new tokens 16 & Single NVIDIA A100 GPU; BF16 inference; single-GPU runtime. \\
mPLUG-Owl3-7B & \texttt{mPLUG/mPLUG-Owl3-7B-241101} & Transformers/ModelScope; sampled frames & Greedy decoding; max new tokens 16 & Single NVIDIA A100 GPU; BF16 inference; single-GPU runtime. \\
LLaVA-Video-7B-Qwen2 & \texttt{lmms-lab/LLaVA-Video-7B-Qwen2} & LLaVA-NeXT checkpoint; sampled frames & Greedy decoding; max new tokens 16 & Single NVIDIA A100 GPU; BF16 inference; single-GPU runtime. \\
Kimi-VL-A3B & \texttt{moonshotai/Kimi-VL-A3B-Instruct} & Transformers checkpoint; sampled frames & Greedy decoding; max new tokens 16 & Single NVIDIA A100 GPU; BF16 inference; single-GPU runtime. \\
Qwen3-VL-30B-A3B & \texttt{Qwen/Qwen3-VL-30B-A3B-Instruct} & Hosted/vLLM-compatible endpoint & Greedy decoding; max new tokens 16 & Hosted open-weight serving; hardware not disclosed. \\
InternVL3.5-30B-A3B & \texttt{OpenGVLab/InternVL3\_5-30B-A3B} & Hosted checkpoint endpoint & Greedy decoding; max new tokens 16 & Hosted open-weight serving; hardware not disclosed. \\
Qwen3-VL-235B-A22B & \texttt{Qwen/Qwen3-VL-235B-A22B-Instruct} & Hosted/vLLM-compatible endpoint & Greedy decoding; max new tokens 16 & Hosted open-weight serving; hardware not disclosed. \\
InternVL3.5-241B-A28B & \texttt{OpenGVLab/InternVL3\_5-241B-A28B-HF} & Hosted checkpoint endpoint & Greedy decoding; max new tokens 16 & Hosted open-weight serving; hardware not disclosed. \\
UniRS & \texttt{IntelliSensing/UniRS} release code and checkpoint & UniRS code; sampled frames & Greedy decoding; max new tokens 16 & Single NVIDIA A100 GPU; BF16 inference; single-GPU runtime. \\
SkyEyeGPT & \texttt{ZhanYang-nwpu/SkyEyeGPT} & Official code; 16 sampled frames per sensor with sensor and time labels & Greedy decoding; max new tokens 16 & -- \\
EarthDial & \texttt{akshaydudhane/EarthDial\_4B\_MS} & Official code; 16 sampled frames per sensor with sensor and time labels & Greedy decoding; max new tokens 16 & -- \\
\bottomrule
\end{tabular}%
}
\end{table}

\FloatBarrier

\subsection{Prompt Templates}
\label{app:prompt_templates}

All models use the same multiple-choice prompt structure. The input-specific prompts differ only in the visual-input description.

\begin{quote}\small
\textbf{EO-only.} You are given an electro-optical aerial video. Answer the following multiple-choice question based only on the provided video. Select one option from A, B, C, and D.\\
Question: \texttt{[QUESTION]}\\
Options: \texttt{[A/B/C/D]}\\
Answer:
\end{quote}

\begin{quote}\small
\textbf{IR-only.} You are given an infrared aerial video. Answer the following multiple-choice question based only on the provided video. Select one option from A, B, C, and D.\\
Question: \texttt{[QUESTION]}\\
Options: \texttt{[A/B/C/D]}\\
Answer:
\end{quote}

\begin{quote}\small
\textbf{EO+IR.} You are given paired electro-optical and infrared aerial observations of the same event. Answer the following multiple-choice question based on the provided EO and IR evidence. Select one option from A, B, C, and D.\\
Question: \texttt{[QUESTION]}\\
Options: \texttt{[A/B/C/D]}\\
Answer:
\end{quote}

\begin{quote}\small
\textbf{Text-only.} Answer the following multiple-choice question. No visual input is provided. Select one option from A, B, C, and D.\\
Question: \texttt{[QUESTION]}\\
Options: \texttt{[A/B/C/D]}\\
Answer:
\end{quote}

\subsection{Input Formatting, Temporal Sampling, and Parsing}
\label{app:input_format}

For the main evaluation, each video pair is sampled at 16 synchronized event-relative timestamps per modality. EO-only input contains sampled EO observations; IR-only input contains sampled IR observations; EO+IR input contains paired EO and IR observations from matched temporal positions, with modality labels. The shared inference settings are summarized in Table~\ref{tab:model_details}, and model-specific serving records are summarized in Tables~\ref{tab:model_serving_details} and~\ref{tab:model_serving_details_cont}.

Temporal sampling is defined on normalized event time rather than fixed raw frame indices. For each retained EO--IR pair, we sample \(T=16\) uniformly spaced normalized timestamps over the retained paired event interval. EO and IR frames are selected at the nearest available decoded timestamp after event-level alignment; this permits small EO/IR stream-duration differences while preserving the same event-relative temporal positions. EO-only and IR-only inputs contain \(T\) sampled frames from the corresponding stream. EO+IR inputs contain \(T\) paired timestamps, i.e., up to \(2T=32\) frames, ordered as \((\mathrm{EO}_{t_1},\mathrm{IR}_{t_1}),\ldots,(\mathrm{EO}_{t_T},\mathrm{IR}_{t_T})\), with explicit modality labels. All evaluated systems, including API and open-weight models, receive sampled frames rather than full native video files in the main protocol. Frames are passed through each model interface or processor with aspect ratio preserved when resizing is required; no target crops, bounding boxes, masks, tracking metadata, or other visual annotations are supplied.

Table~\ref{tab:frame_counts} lists the exact frame count for each input setting. We match temporal positions across models and input settings rather than total image count: halving the timestamps per sensor in EO+IR would confound sensor availability with temporal coverage, and model-specific default interfaces would expose models to different frame counts and event times. Image and visual-token counts may therefore differ. The three remote-sensing VLMs receive the same 16 timestamps, and only the input formatting changes to match each released interface. For SkyEyeGPT and EarthDial, EO+IR input uses 32 frames ordered as \((\mathrm{EO}_{t_1},\mathrm{IR}_{t_1}),\ldots,(\mathrm{EO}_{t_{16}},\mathrm{IR}_{t_{16}})\) with sensor and time labels.

\begin{table}[t]
\centering
\small
\setlength{\tabcolsep}{6pt}
\caption{Frames per input setting under the default protocol.}
\label{tab:frame_counts}
\begin{tabular}{lccc}
\toprule
Input setting & Timestamps & Frames per sensor & Total frames \\
\midrule
EO-only & 16 & 16 EO & 16 \\
IR-only & 16 & 16 IR & 16 \\
EO+IR & 16 & 16 EO + 16 IR & up to 32 \\
\bottomrule
\end{tabular}
\end{table}

Predictions are parsed deterministically by extracting the first valid answer option from the model response. Valid options are \texttt{A}, \texttt{B}, \texttt{C}, and \texttt{D}. If no valid option is found, the prediction is marked incorrect. API transport failures, timeouts, rate-limit responses, or empty responses are retried up to three times with the same prompt, inputs, and decoding settings. Unresolved failures are recorded as invalid outputs and counted incorrect. Repeated calls are used only to recover transport or service failures; no retry-based majority voting, response-level self-consistency, or best-of-\(n\) selection is used. No external LLM judge is used.

\begin{table}[t]
\centering
\small
\setlength{\tabcolsep}{5pt}
\caption{Prediction parsing rules.}
\label{tab:prediction_parsing}
\begin{tabular}{ll}
\toprule
Model output pattern & Parsed prediction \\
\midrule
\texttt{A} & \texttt{A} \\
\texttt{The answer is B.} & \texttt{B} \\
\texttt{Option C is correct.} & \texttt{C} \\
No valid option found & Incorrect \\
Multiple valid options found & First valid option \\
\bottomrule
\end{tabular}
\end{table}

All locally executed open-weight models were evaluated on NVIDIA A100 GPUs. No model training or finetuning was performed. The reported local experiments consist of sampled-frame inference and deterministic scoring over the fixed RAVEN-Bench evaluation set. Including the main benchmark runs and the reported robustness, frame-budget, and free-form controls, local inference required approximately 250--300 A100 GPU-hours. Provider-hosted API models and hosted open-weight endpoints were evaluated through their serving interfaces; their underlying hardware is not disclosed by the providers.

\section{Validity and Robustness Controls}
\label{app:robustness_controls}

This appendix reports controls used to test whether RAVEN-Bench is visually grounded and whether paired-input gains reflect aligned EO--IR evidence rather than answer priors, prompt artifacts, input-budget effects, or temporal-format artifacts.

\subsection{Non-visual Controls and Human Reference}
\label{app:validity_controls}

The random baseline samples uniformly from the four answer choices. The answer-prior control predicts from the marginal answer distribution and is used only as a diagnostic estimate of option-prior leakage, not as a deployable baseline. The text-only setting removes visual input and provides only the question and answer options. The human-reference row is used only as contextual answerability calibration, and the independent audit statistics in Appendix~\ref{app:dataset_versions} summarize annotation reliability.

The random and answer-prior rows in Table~\ref{tab:validity_controls} are computed from fixed-seed sampled control prediction files on the exported evaluation prompts, not from analytic expectations. The random file samples one option per item uniformly from \texttt{A}--\texttt{D}; the answer-prior file samples one option per item from the marginal answer distribution of the exported prompt option order. \(S_{\mathrm{main}}\), \(S_{\mathrm{cons}}\), and \(S_{\mathrm{coh}}\) are then computed by the same scorer used for model predictions. Because \(S_{\mathrm{cons}}\) and \(S_{\mathrm{coh}}\) are nonlinear group metrics, we do not substitute expected question-level accuracy into the group-metric formulas.

The human-reference subset contains 120 QA items from 10 EO--IR video pairs, stratified by scenario family, modality requirement, and capability level. The full audit in Appendix~\ref{app:dataset_versions} is a separate procedure with different annotator roles, instructions, and scoring criteria. The human-reference subset is used as an answerability calibration rather than a model-matched evidence-budget comparison: human annotators may replay the retained EO--IR clips, while models are evaluated under the sampled-frame protocol in Appendix~\ref{app:input_format}. Annotators do not see capability labels, modality requirements, group identifiers, or answer keys. Majority vote determines the final response, with expert adjudication for ties or uncertainty.

The best-MLLM row in Table~\ref{tab:validity_controls} is Gemini~3.1~Pro, selected by \(S_{\mathrm{main}}\) under EO+IR input on the same 120-item subset. The grouped scores in that row are from the same model, not metric-wise maxima over different models.

\begin{table}[t]
\centering
\small
\setlength{\tabcolsep}{4.5pt}
\caption{Validity controls. Scores are percentages. The text-only controls estimate language-prior leakage; the human-reference scores are computed on the stratified 120-item subset with retained EO--IR clips for answerability calibration.}
\label{tab:validity_controls}
\begin{tabular}{lcccc}
\toprule
Setting & Visual input & $S_{\mathrm{main}}$ & $S_{\mathrm{cons}}$ & $S_{\mathrm{coh}}$ \\
\midrule
Random & None & 24.8 & 10.4 & 8.5 \\
Answer-prior & None & 26.1 & 10.8 & 8.8 \\
Text-only (Qwen3.5-9B~\citep{qwen2026qwen35}) & None & 22.9 & 8.3 & 6.6 \\
Text-only (best over evaluated models) & None & 28.4 & 12.1 & 9.8 \\
Human subset (120 QA / 10 pairs) & EO+IR & 82.3 & 77.1 & 69.6 \\
Best MLLM on same subset & EO+IR & 50.4 & 38.6 & 31.8 \\
\bottomrule
\end{tabular}
\end{table}

\subsection{Cross-modal Robustness Controls}
\label{app:gcm_robustness}

For a control condition \(c\), we define
\begin{equation}
G^{c}_{\mathrm{ctrl}}=S^{c}_{\mathrm{main}}-\max\left(S^{\mathrm{EO}}_{\mathrm{main}},S^{\mathrm{IR}}_{\mathrm{main}}\right).
\end{equation}
For the aligned EO+IR condition, this equals \(G_{\mathrm{cm}}\). For duplicated, shuffled, or misaligned inputs, it is used only to test whether the aligned EO+IR improvement can be explained by input-budget or formatting artifacts.

Table~\ref{tab:gcm_robustness} uses model-level aggregation. For each model in the representative subset, \(G^{c}_{\mathrm{ctrl}}\) is computed using that model's control score \(S_{\mathrm{main}}^{c}\) and that same model's EO-only and IR-only scores on the same retained 1,200-item evaluation set. The table then reports the macro-average over models. The representative subset is the same subset used in Figure~\ref{fig:modality_slices}: Gemini~3.1~Pro, Qwen3-VL-235B, Qwen3.5-9B, InternVideo2.5-Chat-8B, UniRS, and LLaVA-Video-7B-Qwen2. Thus \(G^{c}_{\mathrm{ctrl}}\) is not formed by subtracting averaged EO and IR baselines after aggregation.

\begin{table*}[t]
\centering
\small
\setlength{\tabcolsep}{4pt}
\caption{Robustness controls for cross-modal gain. Scores are averaged over a representative model subset.}
\label{tab:gcm_robustness}
\begin{tabular}{lccccp{0.34\linewidth}}
\toprule
Condition & $S_{\mathrm{main}}^{c}$ & $G_{\mathrm{ctrl}}^{c}$ & $S_{L2}$ & $S_{\mathrm{coh}}$ & Control purpose \\
\midrule
Aligned EO+IR & 43.5 & +3.1 & 45.7 & 23.4 & Main paired-input setting. \\
EO+EO & 41.1 & +0.7 & 42.8 & 20.6 & Controls for two-stream input budget using duplicated EO evidence. \\
IR+IR & 40.7 & +0.3 & 43.1 & 21.1 & Controls for two-stream input budget using duplicated IR evidence. \\
Modality-shuffled EO+IR & 39.6 & -0.8 & 41.2 & 18.3 & Tests whether true EO--IR correspondence is required. \\
Temporally misaligned EO+IR & 40.1 & -0.3 & 39.9 & 17.4 & Tests whether synchronized EO--IR timestamps are required. \\
Temporal-shuffle & 39.2 & -1.2 & 37.4 & 15.8 & Tests whether ordered temporal evidence is required. \\
EO--IR label-swap & 40.5 & +0.1 & 41.3 & 19.2 & Tests whether modality labels and sensor semantics are used correctly. \\
Side-by-side mosaic & 39.0 & -1.4 & 39.1 & 17.1 & Tests spatial concatenation and target-size reduction. \\
\bottomrule
\end{tabular}
\end{table*}

The control inputs are constructed from the same sampled-frame protocol as Appendix~\ref{app:input_format}. EO+EO and IR+IR duplicate the corresponding stream into both two-stream input slots, preserving the paired-input frame budget while removing cross-spectral complementarity. Modality-shuffled EO+IR pairs an EO stream with an IR stream from a different retained event, preserving the two-modality format but breaking event correspondence. Temporally misaligned EO+IR preserves the same EO--IR event pair but shifts one modality's timestamp sequence relative to the other. Temporal-shuffle keeps the same sampled frames but permutes their temporal order. EO--IR label-swap exchanges the modality labels while keeping the frames fixed, and side-by-side mosaic concatenates matched EO and IR frames into a single visual frame per timestamp. The aligned EO+IR setting yields the largest control gain and strongest coherence. Duplicated-stream controls preserve the two-stream input budget but produce smaller gains, indicating that paired-input improvement is not explained by additional visual tokens alone. Shuffled, misaligned, and temporal-order controls reduce main, temporal, and coherence scores, supporting the need for aligned cross-spectral and temporal evidence.

\subsection{Frame-budget Sensitivity}
\label{app:frame_budget}

For each frame budget \(T\), EO-only, IR-only, and EO+IR inputs are all resampled at the same \(T\) synchronized timestamps. \(G_{\mathrm{cm}}\) is recomputed from the corresponding EO-only and IR-only baselines at that same \(T\), rather than using the default \(T=16\) unimodal baselines. Scores average the representative six-model subset of Table~\ref{tab:gcm_robustness}.

\begin{table}[t]
\centering
\small
\setlength{\tabcolsep}{5pt}
\caption{Frame-budget sensitivity under paired EO+IR input. \(T\) denotes the number of synchronized timestamps per modality.}
\label{tab:frame_budget}
\begin{tabular}{cccccc}
\toprule
$T$ & $S^{\mathrm{EO+IR}}$ & $G_{\mathrm{cm}}$ & $S_{L2}$ & $S_{\mathrm{coh}}$ & $S^{\mathrm{EO+IR}}-S_{\mathrm{coh}}$ \\
\midrule
4  & 38.4 & +0.9 & 39.6 & 16.8 & 21.6 \\
8  & 41.3 & +2.1 & 43.1 & 20.1 & 21.2 \\
16 & 43.5 & +3.1 & 45.7 & 23.4 & 20.1 \\
32 & 44.1 & +3.2 & 46.3 & 23.9 & 20.2 \\
\bottomrule
\end{tabular}
\end{table}

Increasing the number of synchronized timestamps improves temporal and coherence scores up to a moderate budget, after which gains become smaller. This supports the default sampling budget while showing that the main diagnostic trends are not caused by a single arbitrary frame count. From \(T=16\) to \(T=32\), Main changes by 0.6 points, \(G_{\mathrm{cm}}\) by 0.1, \(S_{L2}\) by 0.6, and Coherence by 0.5, and the Main--Coherence gap is nearly unchanged (20.1 versus 20.2).

\subsection{Free-form Evaluation}
\label{app:free_form}

Multiple-choice questions can make an item easier because the options narrow the answer space; their advantage is deterministic scoring without a language-model judge. To test generation directly, we also ran an open-ended evaluation with a six-model subset (Gemini~3.1~Pro, Qwen3-VL-235B, Qwen3.5-9B, InternVideo2.5-Chat-8B, UniRS, and LLaVA-Video-7B-Qwen2) under the same \(T=16\) protocol.

Every free-form item is derived from one benchmark item. Questions that depended on their options were rewritten to be self-contained, keeping the same target and evidence requirements, time interval, and answer detail. Each model returns a concise answer followed by a brief evidence statement that identifies the required EO and/or IR evidence, when it appears, and how it changes over time. Two annotators, blind to model identity, independently judge answer and evidence correctness with a question-specific rubric; a third annotator resolves disagreements. Agreement is 93.1\% for answer correctness and 88.9\% for evidence correctness (Cohen's \(\kappa=0.85\) and \(0.77\)).

Free-form Main and Answer+Evidence use the same balanced average over L1, L2, and L3 as the multiple-choice Main score. Answer+Evidence requires both judgments to be correct. Free-form Consistency and Coherence are computed from answer correctness only, using the original question groups.

\begin{table}[t]
\centering
\small
\setlength{\tabcolsep}{4pt}
\caption{Free-form evaluation on the 100-pair release, averaged over the six-model subset. MC: multiple-choice; FF: free-form.}
\label{tab:free_form}
\begin{tabular}{lccccc}
\toprule
Evaluation set & MC Main & FF Main & Answer+Evidence & FF Cons. & FF Coh. \\
\midrule
RAVEN-Bench (100 pairs) & 43.5 & 34.8 & 27.1 & 24.6 & 19.3 \\
\bottomrule
\end{tabular}
\end{table}

Free-form Main is 34.8, compared with 43.5 under multiple choice. Question wording, output format, and scoring also change, so the gap reflects more than the removal of answer options. For all six models, free-form Main remains above both grouped scores, and Answer+Evidence is 7.7 points below free-form Main, so some correct answers still lack correct supporting evidence.

\FloatBarrier

\section{Qualitative Error Cases and Responsible Release}
\label{app:failure_release}

This appendix provides qualitative examples for the error mechanisms summarized in
Section~\ref{sec:failure_modes} and documents the responsible-release policy for
RAVEN-Bench.

\subsection{Representative Error Cases}
\label{app:failure_cases}

\begin{figure}[!htbp]
    \centering
    \includegraphics[width=\textwidth]{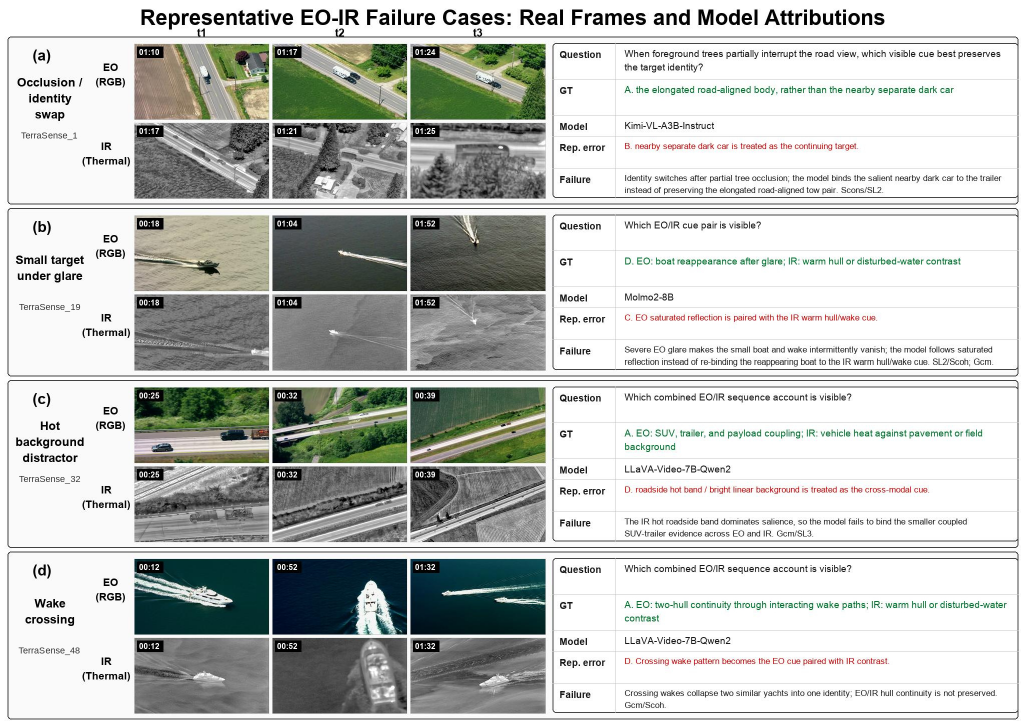}
    \caption{Representative EO--IR failure cases using real sampled frames and representative model outputs. Each row shows temporally aligned EO and IR evidence, the model-facing question, ground-truth answer, model prediction, and a manual diagnostic attribution. The rows illustrate identity swaps under partial occlusion, small-target evidence loss under glare, hot-background thermal distractors, and wake-induced cross-modal binding failures.}
    \label{fig:failure_cases_app}
\end{figure}

Figure~\ref{fig:failure_cases_app} illustrates how visually plausible answers can fail to preserve the evidence needed for EO--IR reasoning. These examples are not intended as a model leaderboard; they are selected to make the observed failure mechanisms concrete. In row (a), foreground-tree occlusion and a nearby road vehicle create an identity-swap error: the model follows a salient nearby dark car instead of maintaining the road-aligned tow pair through the interruption. In row (b), glare and surface reflections reduce the visibility of a small target, causing the model to pair an unstable EO cue with the wrong IR evidence. In row (c), a hot background structure dominates the thermal scene and pulls the model away from the smaller vehicle cue. In row (d), crossing wakes and similar vessel appearances create a cross-modal binding error: the EO wake cue is not reliably associated with the same target in IR.

These cases support the diagnostic role of the grouped metrics. Identity swaps and small-target losses primarily reduce stable target interpretation and ordered evidence preservation, which are exposed by $S_{\mathrm{cons}}$ and $S_{\mathrm{coh}}$. Hot-background and wake-based distractors affect whether paired EO+IR input is used to bind the correct evidence, which is reflected in $G_{\mathrm{cm}}$ and Level-3 performance. The examples therefore explain why aggregate multiple-choice accuracy can overstate evidence-supported EO--IR reasoning.

\FloatBarrier

\subsection{Error-Mechanism Taxonomy}
\label{app:failure_taxonomy}

Table~\ref{tab:failure_taxonomy} groups the qualitative cases into recurring error mechanisms tied to aerial EO--IR sensing conditions. The taxonomy connects observable conditions, model behavior, and the diagnostic readouts most directly affected by each mechanism. The categories are not mutually exclusive: a single failure may involve occlusion, target scale, thermal salience, and cross-modal ambiguity at the same time. The table is therefore intended as a diagnostic aid rather than an exhaustive causal classification.

\begin{table}[!htbp]
\centering
\footnotesize
\setlength{\tabcolsep}{3.5pt}
\renewcommand{\arraystretch}{1.08}
\caption{Remote-sensing error-mechanism taxonomy. Each mechanism links an EO--IR sensing condition to a typical model behavior and the diagnostic readouts most directly affected.}
\label{tab:failure_taxonomy}
\begin{tabular}{@{}p{0.20\linewidth}p{0.28\linewidth}p{0.32\linewidth}p{0.13\linewidth}@{}}
\toprule
Error mechanism & Physical condition & Model behavior & Primary readouts \\
\midrule
Occlusion-driven identity swap
& Foreground trees, moving occluders, nearby vehicles, viewpoint changes, or scale changes interrupt the target track.
& The model binds post-occlusion evidence to a visually similar nearby object instead of preserving the same target identity.
& $S_{\mathrm{cons}}$, $S_{\mathrm{coh}}$ \\
\midrule
Tiny-target evidence loss
& The target occupies only a few pixels after resizing, frame sampling, or video-token compression; glare, reflection, or surface texture further reduces usable evidence.
& The model detects a local cue or salient artifact but fails to preserve the target across frames or modalities.
& $S_{L2}$, $S_{\mathrm{coh}}$ \\
\midrule
Sensor-salience distractor
& EO glare, shadows, wakes, hot backgrounds, low thermal contrast, or thermal crossover dominates the visible or thermal salience field.
& The model follows the visually or thermally dominant distractor rather than the smaller target cue.
& $G_{\mathrm{cm}}$, $S_{L3}$ \\
\midrule
Cross-modal binding failure
& EO and IR provide asymmetric evidence: EO may preserve shape, color, texture, or wake geometry, while IR may emphasize thermal contrast or suppress visible-spectrum cues.
& The model relies on one stream, switches between inconsistent candidates, or pairs an EO cue with the wrong IR evidence.
& $G_{\mathrm{cm}}$, $S_{\mathrm{coh}}$ \\
\bottomrule
\end{tabular}
\end{table}

This taxonomy clarifies why the same aggregate accuracy can correspond to different evidence-use failures. $S_{\mathrm{cons}}$ exposes fragmented interpretation of the same target across related questions. $S_{\mathrm{coh}}$ exposes event-level answers that are not supported by the ordered intermediate evidence. $G_{\mathrm{cm}}$ measures paired-input gain beyond the stronger unimodal stream, and should be interpreted as a controlled diagnostic of EO+IR benefit rather than as direct proof of sensor fusion.

\FloatBarrier

\subsection{Responsible Release and Limitations}
\label{app:responsible_release}

\paragraph{Intended use and license terms.}
RAVEN-Bench is intended for research evaluation and auditing of EO--IR video reasoning. It is not designed for operational deployment or for any setting in which model outputs would inform real-world observation, intervention, or enforcement decisions. 
The videos and annotations are released under the TerraSense Analytics Data Release Terms, which grant a non-exclusive license for research and non-commercial use and require attribution to TerraSense Analytics Ltd. The terms prohibit commercial use of the dataset; attempts to re-identify individuals; unlawful or harmful use, including surveillance, tracking, or profiling of specific individuals, generation of deceptive media, discriminatory profiling, and fully automated decisions that adversely affect legal rights; and use of the licensor's name to suggest endorsement. Users who redistribute the dataset or a modified version must pass these terms on to downstream recipients and identify the modifications. The terms place no restrictions on models or other results derived from the data, and the license terminates if a user breaches it.

\paragraph{What the release contains and omits.}
RAVEN-Bench provides no annotations or metrics for detection, tracking, segmentation, geolocation, targeting, or pixel-level registration. Users with access could still create new annotations from the videos. The release contains sanitized EO and IR benchmark segments, QA annotations, modality requirements, evidence intervals, group metadata, and evaluation code. It contains no boxes, masks, crops, target tracks, trajectories, geolocation, flight paths, pixel-registration data, or KLV and other operational metadata. Temporal annotations mark evidence intervals only. The original collection files are not released.

\paragraph{Residual re-purposing risk.}
Framing RAVEN-Bench as a video-QA benchmark does not prevent other technical uses. Removing operational metadata lowers the risk of exposing collection details, but the visual and temporal content can still be reused: after receiving access, a user could create labels or pseudo-labels and apply the clips to detection, tracking, or other computer-vision tasks beyond the benchmark's intended use. Misuse could include surveillance, targeting, enforcement, border-monitoring, or military applications. Because the EO and IR streams are paired in time but carry no spatial registration, boxes, masks, or correspondence labels, such reuse would require new annotation, and the size of the release limits its training value; it could still support small-scale evaluation or fine-tuning. The data release terms permit non-commercial derivative research and do not restrict derived models, so the remaining safeguards are the prohibitions above, the omitted metadata and spatial labels, and gated access.
We considered releasing only a fixed set of sparse frames, which would reduce some re-purposing risk, especially for tracking and fine-tuning. It would also remove the temporal evidence needed to evaluate target continuity, motion interpretation, and evidence order, and to reproduce the sampling protocol, the frame-budget comparison, and the temporal-order and misalignment controls. We therefore release sanitized segments that preserve this temporal evidence.

\paragraph{Access process.}
The dataset is hosted on Hugging Face (\url{https://huggingface.co/datasets/TerraSenseAnalytics/RAVEN-Bench}) with gated access. The dataset card, with documentation and access instructions, is public. The data files become available after an authenticated user states an institutional affiliation and proposed research use, accepts the data release terms, and the maintainers approve the request. The evaluation code is released under the MIT License at \url{https://github.com/yuhu990424/RAVEN-Bench}. 
These controls cannot prevent every misuse after download. They prevent anonymous access, identify who received the files, and make the license terms a condition of access.

\paragraph{Data ownership and privacy screening.}
The footage was collected by and remains under the institutional control of TerraSense Analytics Ltd., which owns the dataset and authorized this release with management approval. Several authors are affiliated with TerraSense Analytics.
Authorization from the data owner does not replace privacy review. During curation, the retained EO and IR clips were manually screened for identifiable people and readable identifiers, including names, license plates, vehicle or vessel registration numbers, and other unique markings. A localized identifier was blurred when this could be done without affecting the benchmark evidence; otherwise, the clip or the relevant segment was excluded.
Before release, every retained clip was reviewed again and the final files were inspected after redaction. The release criterion is that no identifiable person or readable unique identifier remains.

\paragraph{Limitations.}
RAVEN-Bench is compact and diagnostic rather than class-balanced at web scale. All videos come from one aircraft, one EO/IR gimbal, and one region in western North America, so the reported scores describe this sensing setup. Weather, altitude, and viewing geometry are recorded per event, but several condition subsets remain too small for reliable robustness comparisons. The benchmark evaluates video-QA reasoning rather than pixel-level registration, detection, tracking, or segmentation. Its conclusions are tied to the retained low-altitude aerial EO--IR clips, the chosen temporal sampling protocol, and the evaluated model interfaces. Extending coverage to other regions, platforms, and sensors will require new data, potentially through additional partners.

\paragraph{Release package and maintenance.}
The release package includes versioned files, evaluation code, validated Croissant metadata with Responsible AI fields, a dataset card, access instructions, issue-reporting procedures, and documented maintenance responsibility. Collection continues weekly; new events pass the same release, privacy, and safety review before the EO--IR correspondence and evidence checks.
Future versions will add harder questions that require both EO and IR evidence, longer temporal intervals, deeper evidence chains, condition-specific results once the subsets are large enough, and, once validated, the experimental SAR stream carried by the same aircraft. The EO-only, IR-only, and EO+IR input conditions, correspondence controls, and grouped metrics will remain fixed so that results stay comparable across versions.

\paragraph{Existing assets and terms.}
The experiments use existing API models, open-weight model checkpoints, and baseline code only for evaluation. The corresponding model papers, reports, checkpoints, or provider documentation are cited in the main text and Appendix~\ref{app:experimental_protocol}. We use each asset under its released license or provider terms of service and do not redistribute third-party model weights, API outputs beyond aggregate benchmark results, or third-party code as part of the dataset release. The RAVEN-Bench videos are released under authorization from the data owner for research evaluation, and no third-party copyrighted video sources are included.


\newpage
\section*{NeurIPS Paper Checklist}

\begin{enumerate}

\item {\bf Claims}
    \item[] Question: Do the main claims made in the abstract and introduction accurately reflect the paper's contributions and scope?
    \item[] Answer: \answerYes{} 
    \item[] Justification: The abstract and introduction describe RAVEN-Bench as a paired EO--IR aerial video QA benchmark with modality-conditioned evaluation, hierarchical capability levels, structured question groups, and representative MLLM evaluation. The claims are limited to benchmark construction and diagnostic evaluation, and do not assert a new model, deployment-ready system, or general coverage beyond the low-altitude EO--IR setting studied in the paper.

\item {\bf Limitations}
    \item[] Question: Does the paper discuss the limitations of the work performed by the authors?
    \item[] Answer: \answerYes{} 
    \item[] Justification: Section 6 (Limitations and responsible use) and Appendix E.3 discuss the limitations of RAVEN-Bench, including its single-aircraft, single-sensor, single-region scope, the small size of condition subsets, its compact diagnostic scale, focus on retained low-altitude EO--IR clips, video-QA formulation rather than pixel-level registration, detection, tracking, or segmentation, and dependence on the chosen temporal sampling protocol and evaluated model interfaces.

\item {\bf Theory assumptions and proofs}
    \item[] Question: For each theoretical result, does the paper provide the full set of assumptions and a complete (and correct) proof?
    \item[] Answer: \answerNA{} 
    \item[] Justification: This paper does not include theoretical results.

    \item {\bf Experimental result reproducibility}
    \item[] Question: Does the paper fully disclose all the information needed to reproduce the main experimental results of the paper to the extent that it affects the main claims and/or conclusions of the paper (regardless of whether the code and data are provided or not)?
    \item[] Answer: \answerYes{} 
    \item[] Justification: Appendix C shows the implementation details and settings of our experiments.

\item {\bf Open access to data and code}
    \item[] Question: Does the paper provide open access to the data and code, with sufficient instructions to faithfully reproduce the main experimental results, as described in supplemental material?
    \item[] Answer: \answerYes{} 
    \item[] Justification: Code: \url{https://github.com/yuhu990424/RAVEN-Bench}. Data: \url{https://huggingface.co/datasets/TerraSenseAnalytics/RAVEN-Bench} (gated access; Appendix E.3).

\item {\bf Experimental setting/details}
    \item[] Question: Does the paper specify all the training and test details (e.g., data splits, hyperparameters, how they were chosen, type of optimizer) necessary to understand the results?
    \item[] Answer: \answerYes{} 
    \item[] Justification: Section 5, appendix C and D shows the implementation details and settings of our experiments.

\item {\bf Experiment statistical significance}
    \item[] Question: Does the paper report error bars suitably and correctly defined or other appropriate information about the statistical significance of the experiments?
    \item[] Answer: \answerYes{} 
    \item[] Justification: The paper discusses statistical uncertainty in Section 4.3 and avoids significance claims based on small isolated score gaps. Appendix B.2 specifies the clustered bootstrap procedure used by the evaluation code: question-level and level-wise metrics are resampled at the EO--IR video-pair level, group metrics are resampled at the group level, and derived quantities are recomputed within each bootstrap draw using 10,000 draws and percentile intervals.

\item {\bf Experiments compute resources}
    \item[] Question: For each experiment, does the paper provide sufficient information on the computer resources (type of compute workers, memory, time of execution) needed to reproduce the experiments?
    \item[] Answer: \answerYes{} 
    \item[] Justification: Appendix C reports the model interfaces, decoding settings, and hardware notes for the evaluated systems. The paper states that locally executed models were run on NVIDIA A100 GPUs, that no training or finetuning was performed, estimates the local inference cost at approximately 250--300 A100 GPU-hours, and notes that provider-hosted hardware is not disclosed.
    
\item {\bf Code of ethics}
    \item[] Question: Does the research conducted in the paper conform, in every respect, with the NeurIPS Code of Ethics \url{https://neurips.cc/public/EthicsGuidelines}?
    \item[] Answer: \answerYes{} 
    \item[] Justification: Nothing in this paper violates the NeurIPS Code of Ethics.

\item {\bf Broader impacts}
    \item[] Question: Does the paper discuss both potential positive societal impacts and negative societal impacts of the work performed?
    \item[] Answer: \answerYes{} 
    \item[] Justification: Section 6 and Appendix E.3 discuss the intended positive use of RAVEN-Bench for research evaluation and auditing of EO--IR video reasoning, and also discuss negative societal risks from misuse in surveillance, targeting, enforcement, border monitoring, military, or critical-infrastructure settings. The paper further describes mitigations including metadata removal, exclusion of sensitive clips, gated research-evaluation access, research-use terms, dataset documentation, and maintenance procedures.
    
\item {\bf Safeguards}
    \item[] Question: Does the paper describe safeguards that have been put in place for responsible release of data or models that have a high risk for misuse (e.g., pre-trained language models, image generators, or scraped datasets)?
    \item[] Answer: \answerYes{} 
    \item[] Justification: Appendix E.3 describes safeguards for the responsible release of RAVEN-Bench. The paper states that the dataset is intended only for research evaluation and auditing, states that it is not designed for operational deployment, licenses it for research and non-commercial use with restrictions on re-identification and harmful uses such as surveillance or tracking of specific individuals, removes or anonymizes sensitive metadata and operational details, excludes clips with personally identifiable information or unnecessary sensitive content, and releases the dataset through gated access under the TerraSense Analytics Data Release Terms with dataset documentation and issue-reporting procedures.

\item {\bf Licenses for existing assets}
    \item[] Question: Are the creators or original owners of assets (e.g., code, data, models), used in the paper, properly credited and are the license and terms of use explicitly mentioned and properly respected?
    \item[] Answer: \answerYes{} 
    \item[] Justification: The paper cites the evaluated API models, open-weight checkpoints, and baseline systems, and Appendix C reports their checkpoint or API identifiers and interfaces. Appendix E.3 states that these existing assets are used under their released licenses or provider terms, are not redistributed as part of RAVEN-Bench, and that the released videos are authorized by the data owner with no third-party copyrighted video sources.

\item {\bf New assets}
    \item[] Question: Are new assets introduced in the paper well documented and is the documentation provided alongside the assets?
    \item[] Answer: \answerYes{} 
    \item[] Justification: The paper introduces RAVEN-Bench as a new benchmark asset, released under the TerraSense Analytics Data Release Terms (research and non-commercial use) with gated access to the video files and with evaluation code under the MIT License (Appendix E.3), and documents its composition, metadata schema, annotation protocol, independent audit, evaluation code, access constraints, limitations, and responsible-release process. The release package includes Croissant metadata, a dataset card, research-use terms, access instructions, issue-reporting procedures, and maintenance information. Code: \url{https://github.com/yuhu990424/RAVEN-Bench}

\item {\bf Crowdsourcing and research with human subjects}
    \item[] Question: For crowdsourcing experiments and research with human subjects, does the paper include the full text of instructions given to participants and screenshots, if applicable, as well as details about compensation (if any)? 
    \item[] Answer: \answerNA{} 
    \item[] Justification: This paper does not involve crowdsourcing nor research with human subjects.

\item {\bf Institutional review board (IRB) approvals or equivalent for research with human subjects}
    \item[] Question: Does the paper describe potential risks incurred by study participants, whether such risks were disclosed to the subjects, and whether Institutional Review Board (IRB) approvals (or an equivalent approval/review based on the requirements of your country or institution) were obtained?
    \item[] Answer: \answerNA{} 
    \item[] Justification: This paper does not involve crowdsourcing nor research with human subjects.

\item {\bf Declaration of LLM usage}
    \item[] Question: Does the paper describe the usage of LLMs if it is an important, original, or non-standard component of the core methods in this research? Note that if the LLM is used only for writing, editing, or formatting purposes and does \emph{not} impact the core methodology, scientific rigor, or originality of the research, declaration is not required.
    \item[] Answer: \answerYes{} 
    \item[] Justification: The paper describes the limited use of LLM-based assistance in the QA construction and annotation workflow. As stated in Section 3.2 and Appendix A.3, LLM assistance was used only for wording refinement, option clarity, distractor review, and ambiguity flagging, and was not used to determine visual facts, answer keys, modality requirements, capability labels, or group assignments.

\end{enumerate}

\end{document}